\documentclass[10pt,twocolumn,letterpaper]{article}
           
\usepackage[pagenumbers]{wacv} 

\usepackage{graphicx}
\usepackage{amsmath}
\usepackage{amssymb}
\usepackage{booktabs}
\usepackage{bbm}
\usepackage{multirow}
\usepackage{dblfnote}

\DFNalwaysdouble
\newcommand{\Lagr}{\mathcal{L}}

\newtheorem{definition}{Definition}

\usepackage[pagebackref,breaklinks,colorlinks]{hyperref}

\newcommand{\PAR}[1]{\noindent{\bf #1}}
\usepackage[capitalize]{cleveref}
\crefname{section}{Sec.}{Secs.}
\Crefname{section}{Section}{Sections}
\Crefname{table}{Table}{Tables}
\crefname{table}{Tab.}{Tabs.}

\begin{document}

\title{OVeNet: Offset Vector Network for Semantic Segmentation}

\author{Stamatis Alexandropoulos\\
Princeton University \thanks{
    \begin{minipage}{0.95\columnwidth}
        This work was done when Stamatis Alexandropoulos was in the National Technical University of Athens.
    \end{minipage}}
\and
Christos Sakaridis\\
ETH Z\"urich\\
\and
Petros Maragos\\
National Technical University of Athens\\
}

\maketitle
\begin{abstract}
    Semantic segmentation is a fundamental task in visual scene understanding. We focus on the supervised setting, where ground-truth semantic annotations are available. Based on knowledge about the high regularity of real-world scenes, we propose a method for improving class predictions by learning to selectively exploit information from neighboring pixels. In particular, our method is based on the prior that for each pixel, there is a seed pixel in its close neighborhood sharing the same prediction with the former. Motivated by this prior, we design a novel two-head network, named Offset Vector Network (OVeNet), which generates both standard semantic predictions and a dense 2D offset vector field indicating the offset from each pixel to the respective seed pixel, which is used to compute an alternative, seed-based semantic prediction. The two predictions are adaptively fused at each pixel using a learnt dense confidence map for the predicted offset vector field. We supervise offset vectors indirectly via optimizing the seed-based prediction and via a novel loss on the confidence map. Compared to the baseline state-of-the-art architectures HRNet and HRNet$+$OCR on which OVeNet is built, the latter achieves significant performance gains on three prominent benchmarks for semantic segmentation, namely Cityscapes, ACDC and ADE20K. Code is available at \url{https://github.com/stamatisalex/OVeNet}.
\end{abstract}

\section{Introduction}
\label{sec:intro}

Semantic segmentation is one of the most central tasks in computer vision. In particular, it is the task of assigning a class to every pixel in a given image. It has lots of applications in a variety of fields, such as autonomous driving \cite{cakir2022semantic,8875923}, robotics \cite{tzelepi2021semantic,balloch2018unbiasing}, and medical image processing \cite{ronneberger2015u,amodio2022cuts}, where pixel-level labeling is critical. 

\begin{figure}
  \centering
  \subfloat[Image]{\includegraphics[width=0.495\linewidth]{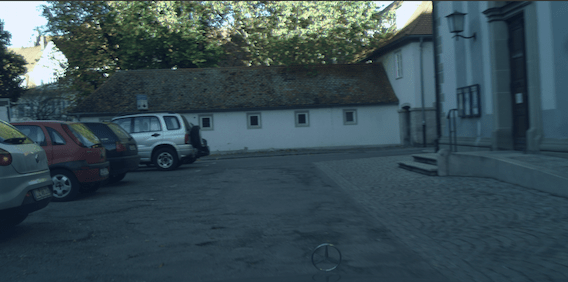}}
  \hfil
  \subfloat[Semantic annotation]{\includegraphics[width=0.495\linewidth]{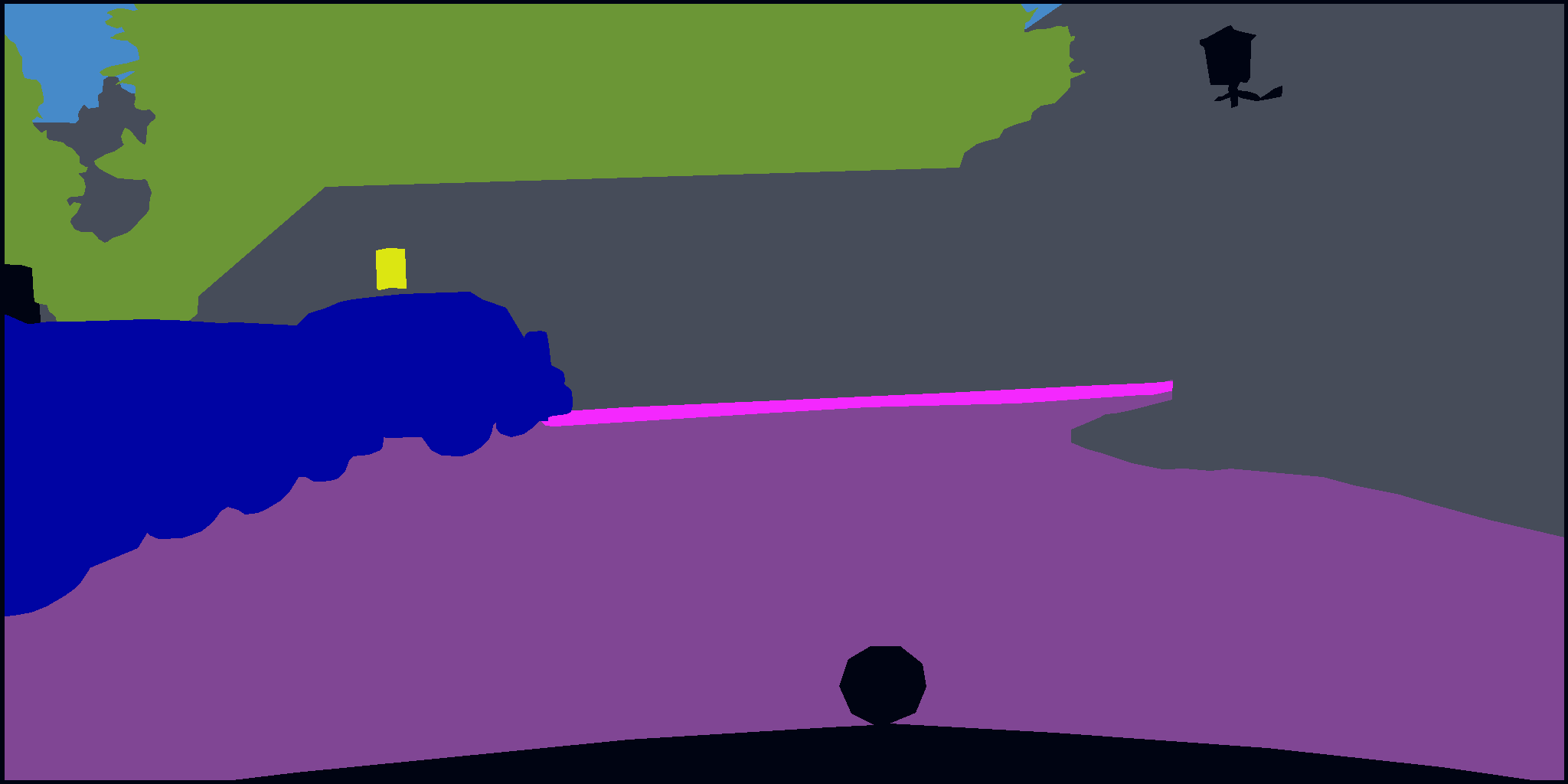}\label{fig:teaser:gt}}
  \\
  \subfloat[HRNet~\cite{wang2020deep} prediction]{\includegraphics[width=0.494\linewidth]{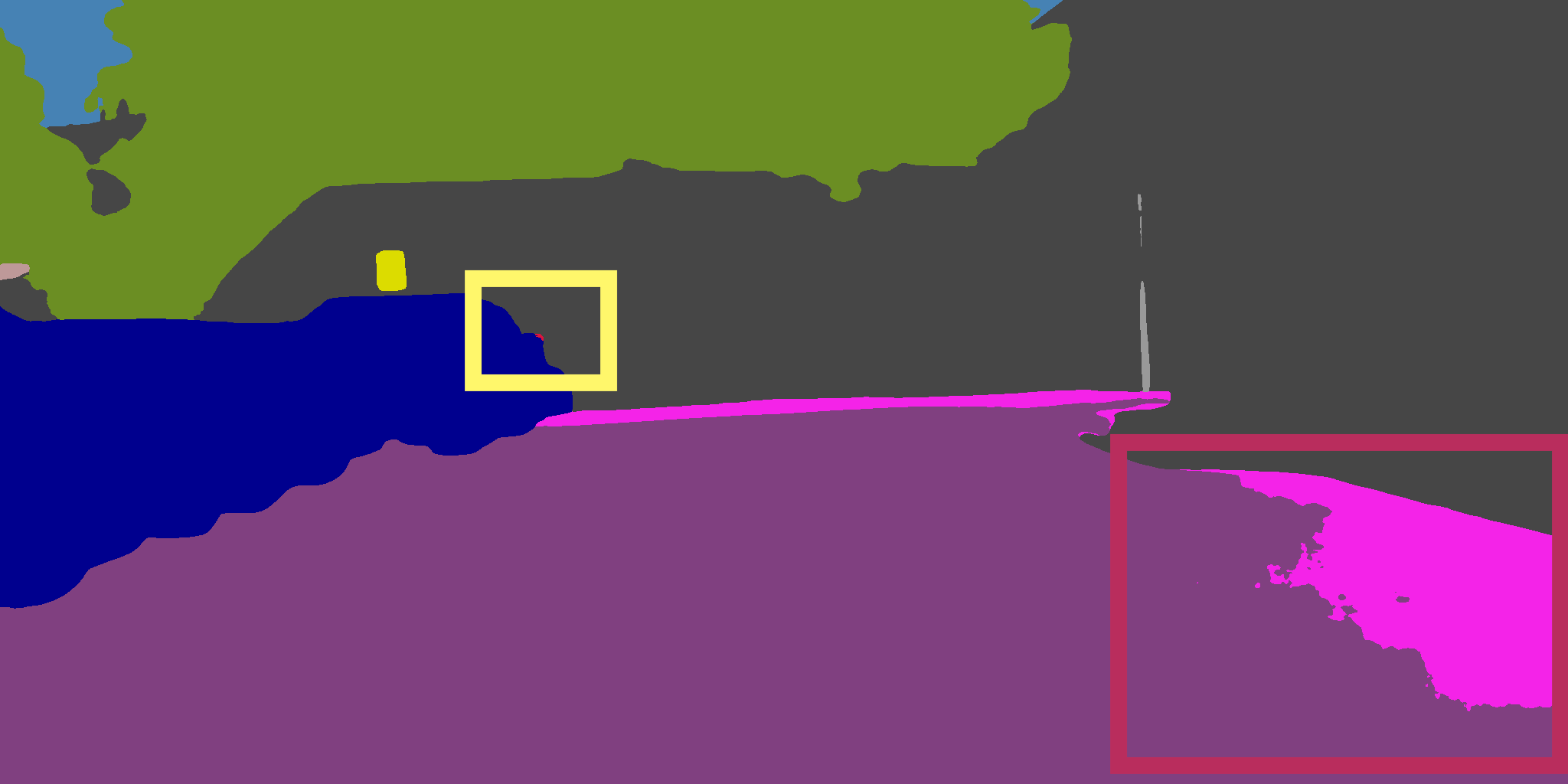}}
  \hfil
  \subfloat[Offset vectors of OVeNet]{\includegraphics[width=0.495\linewidth]{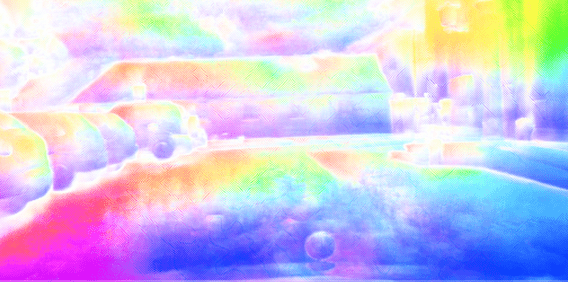}\label{fig:teaser:offset}}

  \subfloat[OVeNet prediction (ours)]{\includegraphics[width=0.495\linewidth]{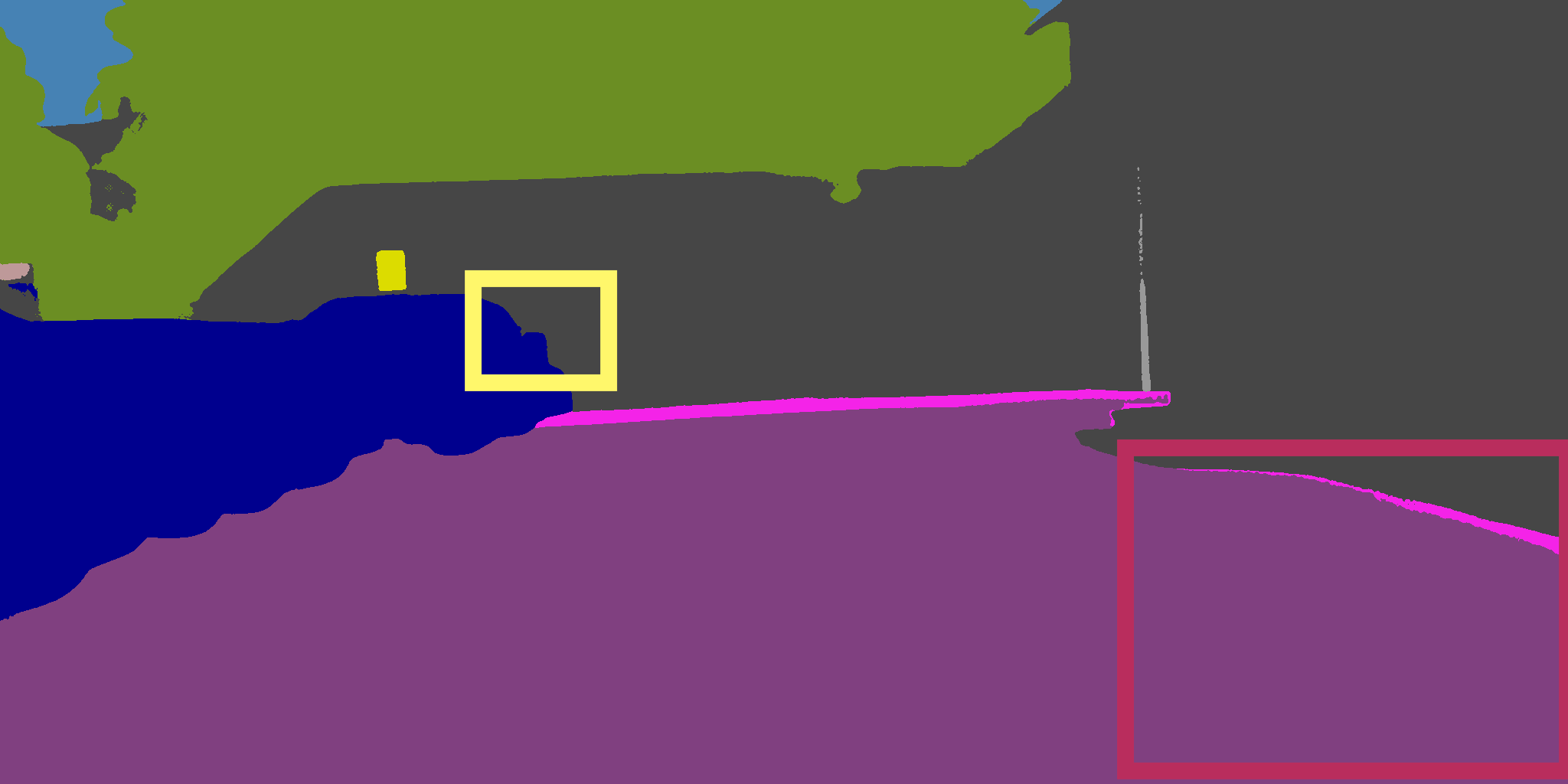}\label{fig:teaser:ours}}
  \caption{The semantic content of real-world scenes has a high degree of regularity. We propose a semantic segmentation method which can exploit this regularity, by learning to selectively leverage information from neighboring \emph{seed} pixels. Our proposed Offset Vector Network (OVeNet) can improve upon state-of-the-art architectures~\cite{wang2020deep} by estimating the offset vectors to such seed pixels and using them to refine semantic predictions.}
  \label{fig:teaser}
\end{figure}

The adoption of convolutional neural networks (CNNs) \cite{long2015fully} for semantic image segmentation has led to a tremendous improvement in performance on challenging, large-scale datasets, such as Cityscapes\cite{cordts2016cityscapes}, MS COCO \cite{lin2014microsoft} and ACDC \cite{sakaridis2021acdc}. Most of the related works \cite{chen2018encoder, wang2020deep, DBLP:journals/corr/YuK15, noh2015learning, chen2017deeplab} focus primarily on architectural modifications of the employed networks in order to better combine global context aggregation and local detail preservation, and use a simple loss that is computed on individual pixels. The design of more sophisticated losses \cite{ke2018adaptive, yuan2019segmentation, neven2019instance} that take into account the structure which is present in semantic labelings has received significantly less attention. Many supervised techniques utilize a pixel-level loss function that handles predictions for individual pixels independently of each other. By doing so, they ignore the high regularity of real-world scenes, which can eventually profit the final model's performance by leveraging information from adjacent pixels. Thus, these methods misclassify several pixels, primarily near semantic boundaries, which leads to major losses in performance.

Based on knowledge about the high regularity of real scenes, we propose a method for improving class predictions by learning to selectively exploit information from neighboring pixels. In particular, the general nature of this idea is applicable on SOTA models like HRNet \cite{wang2020deep} or HRNet $+$ OCR \cite{yuan2019segmentation} and can extend these models by adding a second head to them capable of achieving this goal.
 
The architecture of our Offset Vector Network (OVeNet) is shown in Fig.~\ref{fig:method}. In particular, following the base architecture of the backbone model (e.g HRNet or HRNet $+$ OCR), the first head of our network outputs the initial pixel-level semantic predictions. In general, two pixels $\mathbf{p}$ and $\mathbf{q}$ that belong to the same category share the same semantic outcome. If the pixels belong to the same class, using the label of $\mathbf{q}$ for estimating class at the position of $\mathbf{p}$ results in a correct prediction. 

We leverage this property by learning to identify \emph{seed pixels} which belong to the same class as the examined pixel, whenever such pixels exist, in order to selectively use the prediction at the former for improving the prediction at the latter. This idea is motivated by a prior which states for every pixel $\mathbf{p}$ associated with a 2D semantically segmented image, there exists a seed pixel $\mathbf{q}$ in the neighborhood of $\mathbf{p}$  which shares the same prediction with the former. In order to predict classes with this scheme, we need to find the regions where the prior is valid. Furthermore, in order to point out the seed pixels in these regions, we must predict the offset vector $\mathbf{o(p)} = \mathbf{q} - \mathbf{p}$ for each pixel $\mathbf{p}$.

As a result, we design a second head that generates a dense offset vector field and a confidence map. The predicted offsets are used to resample the class predictions from the first head and generate a second class prediction. The outcomes from the two heads are fused adaptively using the learnt confidence map as fusion weights, in order to down-weigh the offset-based prediction and rely primarily on the basic class prediction in regions where the prior is not valid. Thanks to using seed pixels for prediction, our network classifies several pixels with incorrect initial predictions, e.g., boundary pixels, to the correct classes. Thus, it improves the shape as well as the form of the corresponding segments, leading to more realistic results. Last but not least, we propose a confidence loss which supervises the confidence map explicitly and further improves performance. An illustrative example of this concept is depicted in Fig.~\ref{fig:teaser}, where OVeNet outperforms the baseline HRNet~\cite{wang2020deep} model, since it enlarges correctly the road and the car segment (red and yellow frame correspondingly) and reduces the total number of misclassified pixels.

We evaluate our method extensively on 3 primary datasets, each serving a specific purpose. For semantic segmentation in driving scenes, we focus on two benchmarks: Cityscapes~\cite{cordts2016cityscapes} and ACDC~\cite{sakaridis2021acdc}. Additionally, we broaden our evaluation by incorporating the ADE20K~\cite{zhou2016semantic,zhou2017scene} dataset, which covers a diverse range of images spanning various indoor and outdoor scenes. We implement our offset vector branch both on HRNet~\cite{wang2020deep} and HRNet$+$OCR~\cite{YuanCW18}. Our approach significantly improves the initial models' output predictions by achieving better mean and per-class results. We conduct a thorough qualitative and quantitative experimental comparison to show the clear advantages of our method over previous SOTA techniques.

\section{Related Work}
\label{sec:formatting}
\PAR{Semantic segmentation architectures.}
Fully convolutional networks~\cite{7478072, sermanet2013overfeat} were the first models that re-architected and fine-tuned classification networks to direct dense prediction of semantic segmentation. They generated low-resolution representations by eliminating the fully-connected layers from a classification network (e.g AlexNet~\cite{krizhevsky2012imagenet}, VGGNet~\cite{simonyan2014very}  or GoogleNet~\cite{szegedy2015going})  and then estimating coarse segmentation maps from those representations. To create medium-resolution representations~\cite{chen2014semantic,li2018detnet,chen2017deeplab, chen2017rethinking, DBLP:journals/corr/YuK15}, fully convolutional networks were expanded using dilated/atrous convolutions, which replaced a few strided convolutions and their associated ones. Following, in order to restore high-resolution representations from low-resolution representations an upsample process was used. This process involved a subnetwork that was symmetric to the downsample process (e.g VGGNet~\cite{simonyan2014very}), and included skipping connections between mirrored layers to transform the pooling indices (e.g. DeconvNet~\cite{noh2015learning}). Other methods include duplicating feature maps, which is used in architectures like U-Net \cite{ronneberger2015u} and Hourglass \cite{NewellYD16, ChuYOMYW17, YangLOLW17, KeCQL18, YangLZ17, DengTZZ17, BulatT17a, TangYW18} or encoder-decoder architectures \cite{7803544,PengFWM16}. Lastly, the process of asymmetric upsampling \cite{BulatT16, ChenWPZYS17, XiaoWW18, LinDGHHB17,ValleBVB18,PengZYLS17, ZhangZPXS18, IslamRBW17} has also been extensively researched. 

The models' representations were then enhanced to include multi-scale contextual information \cite{ZhaoSQWJ17,ChenPKMY18,ChenYWXY16}. PSPNet \cite{zhao2017pyramid} utilized regular convolutions on pyramid pooling representations to capture context at multiple scales, while the DeepLab series \cite{chen2017rethinking, chen2017deeplab} used parallel dilated convolutions with different dilation rates to capture context from different scales. Recent research \cite{he2019adaptive, yang2018denseaspp, li2016iterative} proposed extensions, such as DenseASPP \cite{yang2018denseaspp}, which increased the density of dilated rates to cover larger scale ranges, or HS3\cite{borse2021hs3}, which supervised intermediate layers in a segmentation network to learn
meaningful representations by varying task complexity. Other studies \cite{lin2019zigzagnet, chen2018encoder,fu2019adaptive} used encoder-decoder structures to exploit the multi-resolution features as the multi-scale context. Here belongs the HRNet \cite{wang2020deep}, the baseline model of our method. HRNet connects high-to-low convolution streams in parallel. It ensures that high-resolution representations are maintained throughout the entire process and creates dependable high-resolution representations with accurate positional information by repeatedly merging the representations from various resolution streams. Applying additionally the OCR \cite{yuan2019segmentation} method, HRNet $+$ OCR is one of the leading models in the task of semantic segmentation.

Lately, transformers have been successful in computer vision tasks demonstrating their effectiveness. ViT \cite{dosovitskiy2020image} was the first attempt to use the vanilla transformer architecture \cite{vaswani2017attention} for image classification without extensive modification.  Unlike later methods,  such as PVT \cite{wang2021pyramid} and Swin \cite{liu2021swin}, that incorporated vision-specific inductive biases into their architectures, the plain ViT suffers inferior performance on dense predictions due to weak prior assumptions. To tackle this problem, the ViT-Adapter \cite{chen2022vision} was introduced, which allowed plain ViT to achieve comparable performance to vision-specific transformers and achieves the SOTA performance on this task.

\PAR{Semantic segmentation loss functions.} Image segmentation has highly correlated outputs among the pixels. Converting pixel labeling problem into an independent problem can lead to problems such as producing results that are spatially inconsistent and have unwanted artifacts, making pixel-level classification unnecessarily challenging. To solve this problem, several techniques \cite{krahenbuhl2011efficient, chen2015learning, zheng2015conditional, liu2015semantic} have been developed, such as integrating structural information into segmentation. For instance, Chen et al. \cite{chen2017deeplab} utilized denseCRF \cite{krahenbuhl2011efficient} for refining the final segmentation result. Following, Zheng et al. \cite{zheng2015conditional} and Liu et al. \cite{liu2015semantic} made the CRF module differentiable within the deep neural network. Other methods that have been used to encode structures include pairwise low-level image cues like grouping affinity (e.g. SPNs~\cite{liu2017learning}, Affinity CNNs~\cite{maire2016affinity}) and contour cues \cite{bertasius2016semantic, chen2016semantic}. InverseForm~\cite{borse2021inverseform}
is another boundary-aware loss term  using an inverse-transformation network, which efficiently learns the degree of parametric transformations between estimated and target boundaries. GANs \cite{radford2015unsupervised} are an alternative for imposing structural regularity in the neural network output. However, these methods may not work well in cases where there are changes in visual appearance or may require expensive iterative inference procedures. Thus, Ke et. al. \cite{ke2018adaptive} introduced AAFs, which are easier to train than GANs and more efficient than CRF without run-time inference. There has been also proposed another loss function \cite{neven2019instance} suitable for in real time applications that pulls the spatial embeddings of pixels belonging to the same instance together.

\PAR{Offset Vector-Based methods} are essential for image analysis tasks that involve adjacent pixels. In particular, they can effectively exploit the information contained in neighbouring pixels, handle image distortions and noise, and improve the accuracy of various image analysis tasks. They can be used in applications such as depth estimation \cite{Patil_2022_CVPR, park2020non} or semantic segmentation \cite{yuan2020segfix, neven2019instance}. Non-local SPNs~\cite{park2020non} enhance depth completion by iteratively refining initial depth predictions using non-local neighbors. Based on knowledge about the high regularity of real 3D scenes,  P3Depth \cite{Patil_2022_CVPR} is another method used for 3D depth estimation that learns to selectively leverage information from coplanar pixels to improve the predicted depth. In semantic segmentation, SegFix \cite{yuan2020segfix} is a model-agnostic post-processing scheme that improved the boundary quality for the segmentation result. Motivated by the empirical observation that the label predictions of interior pixels are more reliable, SegFix replaced the originally unreliable predictions of boundary pixels by the predictions of interior pixels.
OVeNet provides another perspective to use offset vectors and structure modeling by matching the relations between neighbouring pixels in the label space. Although our approach is inspired by the P3Depth idea, we focus on a different task in the 2D world which is semantic segmentation. Moreover, the key difference between our method and SegFix is the timing of when they are applied. Essentially, OVeNet integrates the offset vector learning process into the model training, while SegFix applies the offset correction as a separate post-processing step.


\section{Method}

\begin{figure*}
\begin{center}
\end{center}
    \includegraphics[width=\linewidth]{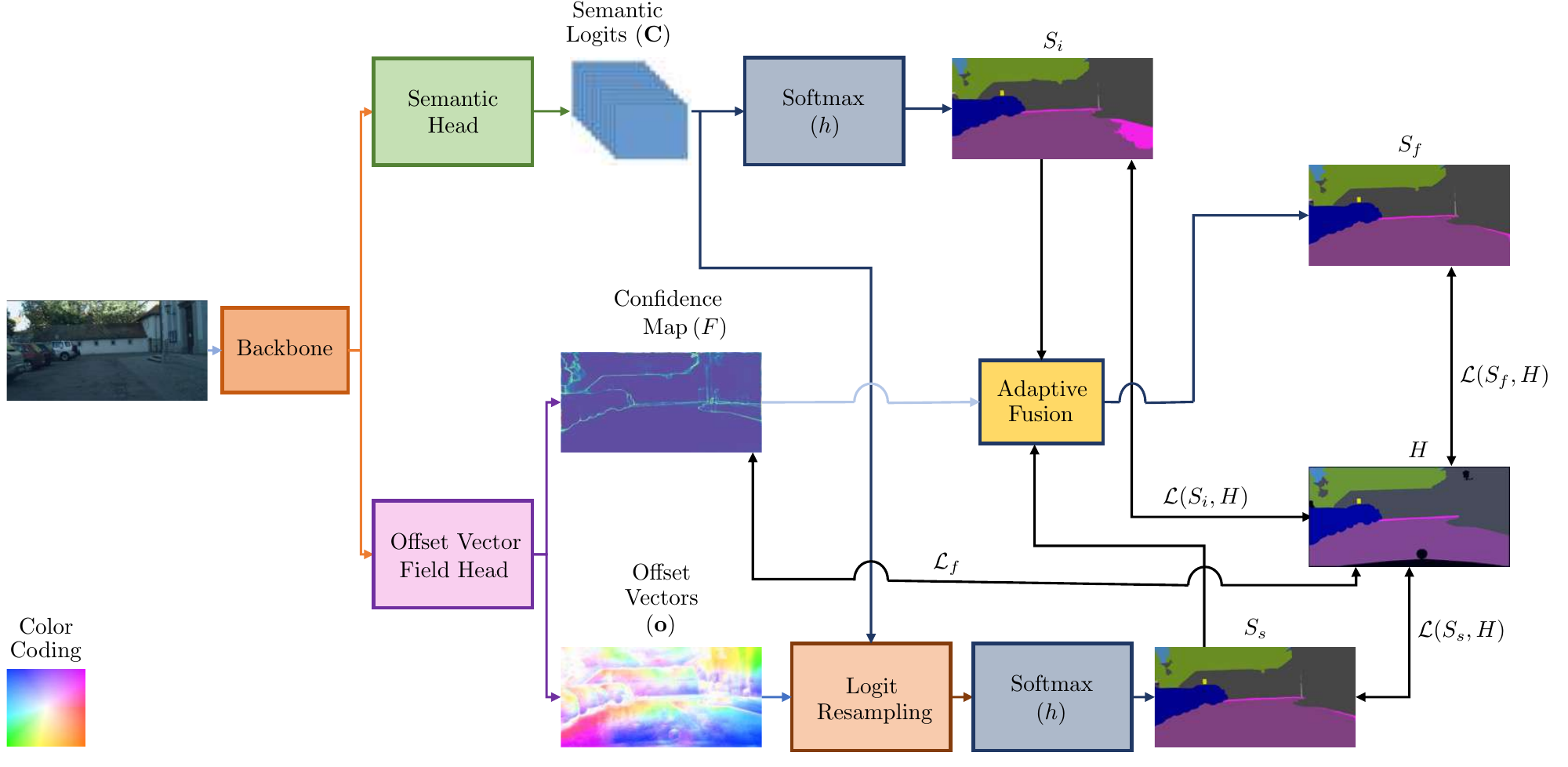}
   \caption{\textbf{Overview of OVeNet.} OVeNet is a two-headed network. The first head outputs semantic logits ($\mathbf{C}$), while the second head outputs a dense offset vector field  ($\mathbf{o}$) identifying positions of seed pixels along with a confidence map ($F$). The logits are passed through a softmax function and output the initial class prediction ($S_i$) of the model. Then, the offset vectors are used to resample the logits from the first head and generate a second class prediction ($S_s$). The two predictions are  adaptively fused using the confidence map resulting in the final prediction $S_f$. For the visualization of the offset vectors we use the optical flow color coding from \cite{ilg2017flownet}. Smaller vectors are lighter and color represents the direction.}
\label{fig:method}
\end{figure*}

In this section, we will analyze our method shown in Fig. \ref{fig:method}. Firstly, in Sec. \ref{notation} we give some basic notation and terminology of semantic segmentation. As we mentioned before in Sec. \ref{sec:intro}, our network estimates semantic labels by selectively combining information from each pixel and its corresponding seed pixel. The intuition and the advantages of using seed pixels to improve the initial prediction of a model are described analytically in Sec. \ref{seed}. Lastly, in Sec. \ref{confidence} we introduce an additional confidence loss, which further enhances our method. 

\subsection{Terminology}
\label{notation}
Semantic segmentation requires learning a dense mapping $f_\theta : I(u,v) \rightarrow S(u,v)$ where  $I$ is the input image with spatial dimensions $H\times W$, $S$ is the corresponding output prediction map of the same resolution, $(u, v)$ are pixel coordinates in the image space and $\theta$ are the parameters of the mapping $f$. In supervised semantic segmentation, a ground-truth semantically segmented map $H$ is available for each image during training. The aim is to optimize the function parameters $\theta$ such that the predicted output map is as close as possible to the ground-truth map across the entire training set $T$. This can be achieved by minimizing the difference between the predicted and ground-truth images:
\begin{equation}
    \label{eq:semantic}
    \min_{\theta} \sum_{(I,H)\epsilon T} \Lagr(f_\theta(I),H)
\end{equation}
where $\Lagr$ is a loss function that penalizes variations between the prediction and the ground truth.

\subsection{Seed Pixel Identification} 
\label{seed}

Let us assume we have one pixel $\mathbf{p}$ which belongs to segment of a semantically segmented image. By definition, every other pixel on this segment has the same class value. Thus, ideally,  in order to get all of the class values accurate, the network only has to predict the class at one of these pixels, $\mathbf{q}$. This pixel can be interpreted as the seed pixel that describes the segment-class. Finally, we let the network find this seed pixel and the corresponding region. 

Let us define a prior which is a relaxed version of the previous idea.

\begin{definition}
For every pixel $\mathbf{p}$ associated with a 2D semantically segmented image, there exists a seed pixel $\mathbf{q}$ in the neighborhood of $\mathbf{p}$ which shares the same prediction with the former.
\label{def:definition}
\end{definition}

In general, there may be numerous seed pixels for $\mathbf{p}$ or none at all.  Given that the Definition \ref{def:definition} holds, semantic segmentation task for $\mathbf{p}$ can be solved by identifying $\mathbf{q}$. For this reason, we let our network predict the offset vector $\mathbf{o(p)} = \mathbf{q} - \mathbf{p}$. Thus, we design our model so that it features a second, offset head and let this offset head predict a dense offset vector field $\mathbf{o}(u, v)$. The two heads of the network share a common main body and then they follow different paths. We resample the initial logits $\mathbf{C}$, being predicted by the first head,  using the estimated offset vector field via: 
\begin{equation}
    \label{eq:seed}
    \mathbf{C_s(p) = C(p + o(p))}
\end{equation}
To manage fractional offsets, bilinear interpolation is used. The resampled logits are then used to compute a second semantic segmentation prediction: 
\begin{gather}
    \label{eq:seed_pred}
    S_s(u,v) = h(\mathbf{C_s}(u,v),u,v) \nonumber \\
    \implies S_s(\mathbf{p}) = S_i(\mathbf{p + o(p)}) 
\end{gather}
based on the seed locations. In our experiment, $h = softmax$.  

Due to the fact that the prior is not always correct, the initial semantic prediction $S_i$ may be preferred to the seed-based prediction $S_s$. To account for such
cases, the second head additionally predicts a confidence map $F(u,v) \in [0,1]$, which represents the model's confidence in adopting the predicted seed pixels for  semantic segmentation via $S_s$. By adaptively fusing $S_i$ and $S_s$, the confidence map is used to compute the final prediction: 
\begin{equation}
    \label{eq:seed_pred_final}
    S_f(\mathbf{p}) = (1-F(\mathbf{p}))S_i(\mathbf{p}) + F(\mathbf{p})S_s(\mathbf{p})
\end{equation}
We apply supervision to each of $S_f$, $S_s$, and $S_i$ in our model, by optimizing the following loss: 
\begin{equation}
    \label{eq:seed_pred_loss}
    \Lagr_{\text{semantic}} = \Lagr(S_f,H) + \kappa\Lagr(S_s,H) + \lambda\Lagr(S_i,H)
\end{equation}
with $\kappa$ and $\lambda$ being hyperparameters and H denoting the GT train ids of each class for each pixel. In this way, we encourage the initial model's head to output an accurate representation across all pixels, even when they have a high confidence value, and the offset vector head to learn high confidence values for pixels for which  Definition~\ref{def:definition} holds and low confidence values for pixels for which the prior does not.

Nonetheless, there is a downside to this approach. Since the model is not supervised directly on the offsets, it has the potential to predict zero offsets across the board. This implies $S_s$ and $S_f$ predictions equivalent to $S_i$. Since the initial predictions $S_i$ are erroneously smoothed around semantic boundaries due to the regularity of the mapping $f_\theta$ in the case of neural networks, this undesirable behavior is avoided in practice. We opt for predicting non-zero offsets that point away from the boundary. Such a non-zero offset utilizes a seed pixel for $S_s$ located further from the border and diminishes inaccuracies stemming from smoothing.  Furthermore, these non-zero offsets extend from the boundaries into the inner sections of regions with smooth segments, aiding the network in forecasting non-trivial offsets, thanks to the regularity of the mapping that forms the offset vector field. Thus, pixels on either side of the boundary have a lower $\Lagr_{\text{semantic}}$ value.

\subsection{Confidence-Based Loss}
\label{confidence}

Our confidence loss is based on the concept that given a pixel coordinate, its surrounding pixels should be in the same segment. For each pixel $\mathbf{p}$, we define the confidence loss as follows:

\begin{align}
    \label{eq:conf_loss}
    \Lagr_{f}(\mathbf{p}) &= -\mathbbm{1}_{H(\mathbf{p})=H(\mathbf{p + o(p)}) }\log F(\mathbf{p}) \nonumber \\
    &\quad - \mathbbm{1}_{H(\mathbf{p})\neq H(\mathbf{p + o(p)})}\log (1-F(\mathbf{p})) 
\end{align}

This idea is motivated by the fact that the confidence value should be large for pixels whose offset vector points to seed pixels with the same class. Similarly, the confidence value should be small for pixels whose offset vector points to seed pixels with a different class. 

To sum up, the complete loss is: 

\begin{equation}
    \label{eq:seed_pred_final_loss}
     \Lagr_{\text{final}}= \Lagr_{\text{semantic}}+ \Lagr_{f}
\end{equation}


\section{Experiments}

To evaluate the proposed method, we carry out comprehensive experiments on the Cityscapes, ACDC and ADE20K datasets. Experimental results demonstrate that our method, compared to the baseline state-of-the-art architectures HRNet~\cite{wang2020deep} and HRNet $+$ OCR~\cite{YuanCW18} on which it is built, achieved higher performance, outperforming these baselines. In the following, we first introduce the datasets, evaluation metrics and implementation details in Sec. \ref{experiment}. We then compare our method to SOTA approaches in Sec. \ref{comparison}. Finally, we perform a series of ablation experiments on Cityscapes in Sec. \ref{ablation}.

\subsection{Experimental Setup}
\label{experiment}

In this section, we present the Cityscapes, ACDC and ADE20K datasets, which are used to evaluate our approach. Evaluation on these datasets is performed using standard semantic segmentation metrics explained below.

\PAR{Cityscapes} \cite{cordts2016cityscapes}  is a challenging urban scene understanding dataset. There are 30 classes from which only 19 classes are used for parsing evaluation. Around 5000 high quality pixel-level finely annotated images and 20000 coarsely annotated images make up the collection. The finely annotated 5000 images are divided into 2975, 500, 1525 images for training, validation and testing respectively.

\PAR{ACDC} \cite{sakaridis2021acdc} is a demanding dataset, used for training and testing semantic segmentation methods on adverse visual conditions. There are 4006 images divided evenly between four frequent unfavorable conditions: fog, dark, rain, and snow and 19 semantic classes, coinciding exactly with the evaluation classes of the Cityscapes dataset. It includes 1600 training and 406 validation images with public annotations and 2000 test images with annotations withheld for benchmarking purposes.

\PAR{ADE20K} \cite{zhou2016semantic} is a challenging scene parsing dataset which covers a diverse range of images depicting various indoor and outdoor scenes. It consists of 20210 images as the training set and 2000 images as the validation set. There are totally 150 semantic classes, including categories like sky, road, grass and discrete objects like person, car, bed.

\PAR{Evaluation Metrics.} The mean of class-wise intersection over union (mIoU) is adopted as the evaluation metric. In addition to the mean of class-wise intersection over union (mIoU), we report other three scores on the test set: IoU category (cat.), iIoU class (cla.) and iIoU category (cat.)

\PAR{Implementation Details.} Our network consists of  two heads. The first head outputs 19 channels, one for each class. The second head outputs three channels: one for each coordinate of the offset vectors and one for confidence. These two heads follow the structure of HRNet. Both OVeNet and the baseline HRNet are initialized with pre-trained ImageNet \cite{krizhevsky2012imagenet} weights. This initialization is important to achieve competitive results as in \cite{wang2020deep}. Following the same training protocol as in \cite{wang2020deep}, the data are augmented by random cropping (from 1024 × 2048 to 512 × 1024 in Cityscapes and from 1080 x 1920 to 540 x 960 in ACDC), random scaling in the range of [0.5, 2], and random horizontal flipping.We use the SGD optimizer with a base learning rate of 0.01, a momentum of 0.9 and a weight decay of 0.0005. The number of epochs used for training is 484. For lowering the learning rate, a poly learning rate policy with a power of 0.9 is applied. The offset vectors are restricted via a tanh layer to have a maximum length of $\tau$ in normalized image coordinates. After an ablation study shown in Sec. \ref{ablation}, we set $\tau$, $\lambda$ and $\mu$ to 0.5 by default and branching is applied at the last ($4^{th}$) stage of HRNet.  The confidence map is predicted through a sigmoid layer. For $S_i$, $S_s$ and $S_f$ predictions, Ohem Cross Entropy Loss\cite{Shrivastava_2016_CVPR} is used. We first experimented with predicting $S_i$, $S_s$ and $S_f$ directly, but this led to inferior results. On the other hand, our final model for which we report performance in the paper outputs the $S_i$, $S_s$, $S_f$ logits. In addition, the confidence based loss is applied using the final semantic prediction $S_f$.

\subsection{Comparison with the State of the Art}
\label{comparison}

\PAR{Cityscapes.} The results on Cityscapes are shown below. We achieve better results on Cityscapes than both the initial HRNet and HRNet $+$ OCR model under similar training time, outperforming prior SOTA methods based on HRNet backbones. Table \ref{tab:test_cityscapes} compares our method with SOTA methods on the Cityscapes test set. All the results are with six scales and flipping. Two cases w/o using coarse data are evaluated: one is about the model learned on the train set, and the other on the train $+$ val set. Our offset vector model excels in both cases with performance gains of $1.4\%$ in mIoU, $2.4\%$ in iIoU cat.\, $0.4\%$ in IoU cat.\ and $1.4\%$ in iIoU cat.\ over the HRNet model  learned only on train set and a gain of $0.5\%$ in mIoU over the HRNet $+$ OCR model learned on both train  $+$ val set. The OVeNet model which is built on HRNet and trained on the training set outperforms the HRNet baseline model which is trained on the training $+$ val set. Table \ref{tab:class} thorougly compares our approach with HRNet’s per-class results. Our method achieves better results in the majority of classes. Our offset vector-based model learns an implicit representation of different objects which can benefit the overall semantic segmentation estimation capability of the network. Regarding val set results, HRNet achieves $81.8\%$ mIoU while OVeNet built on it surpasses it reaching $82.4\%$ mIoU .

\begin{figure*}
  \centering
  \subfloat{\includegraphics[width=0.165\linewidth]{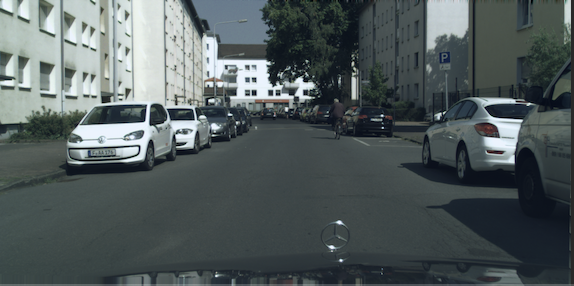}}
  \hfil
  \subfloat{\includegraphics[width=0.165\linewidth]{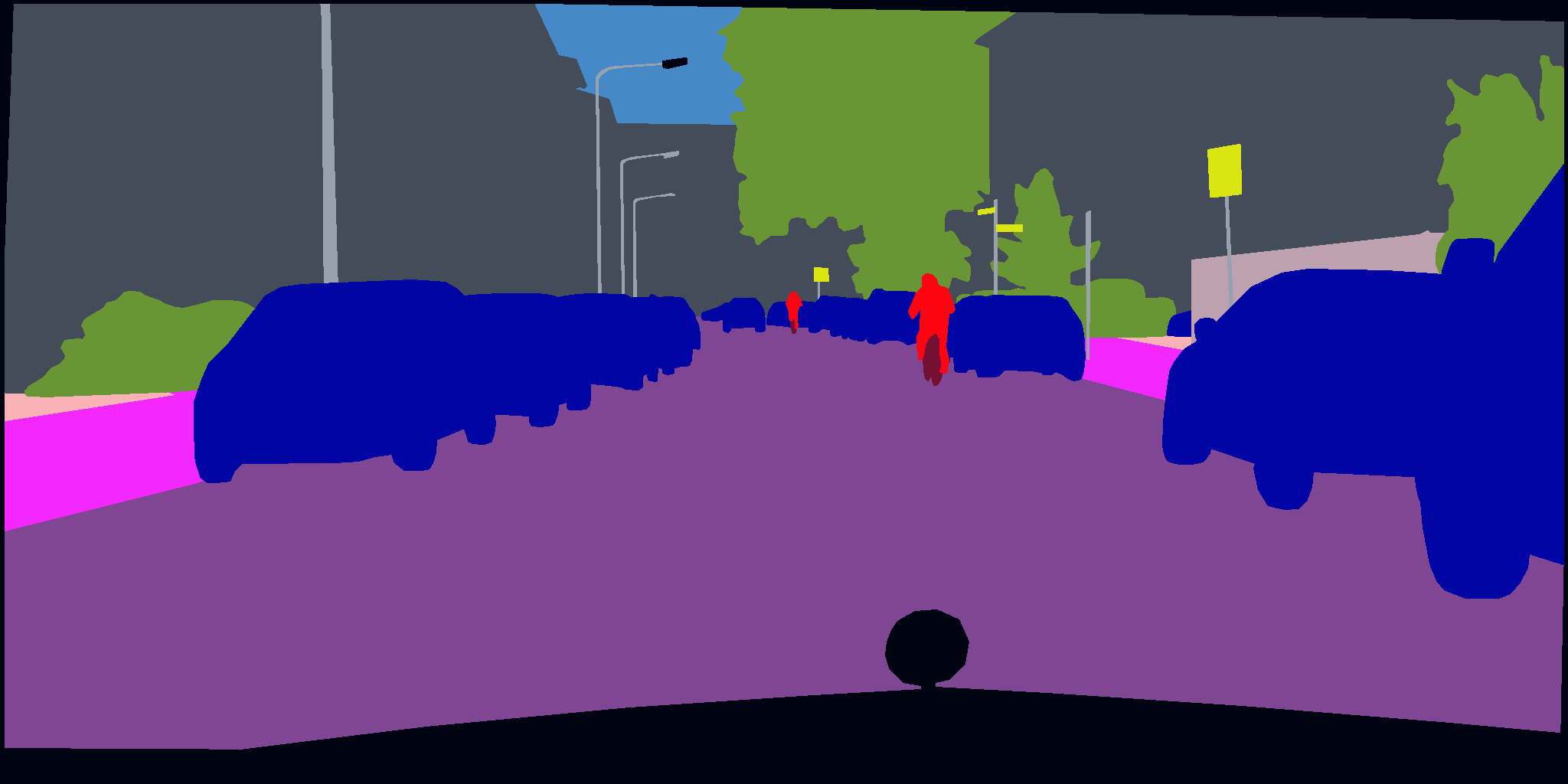}}
  \hfil
  \subfloat{\includegraphics[width=0.165\linewidth]{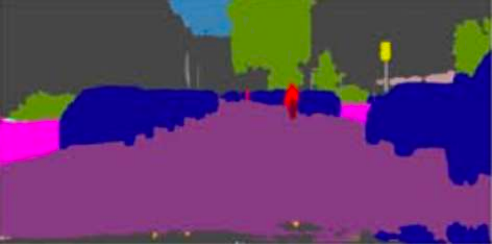}}
  \hfil
  \subfloat{\includegraphics[width=0.165\linewidth]{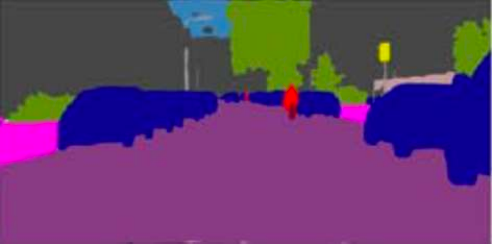}}
  \hfil
  \subfloat{\includegraphics[width=0.165\linewidth]{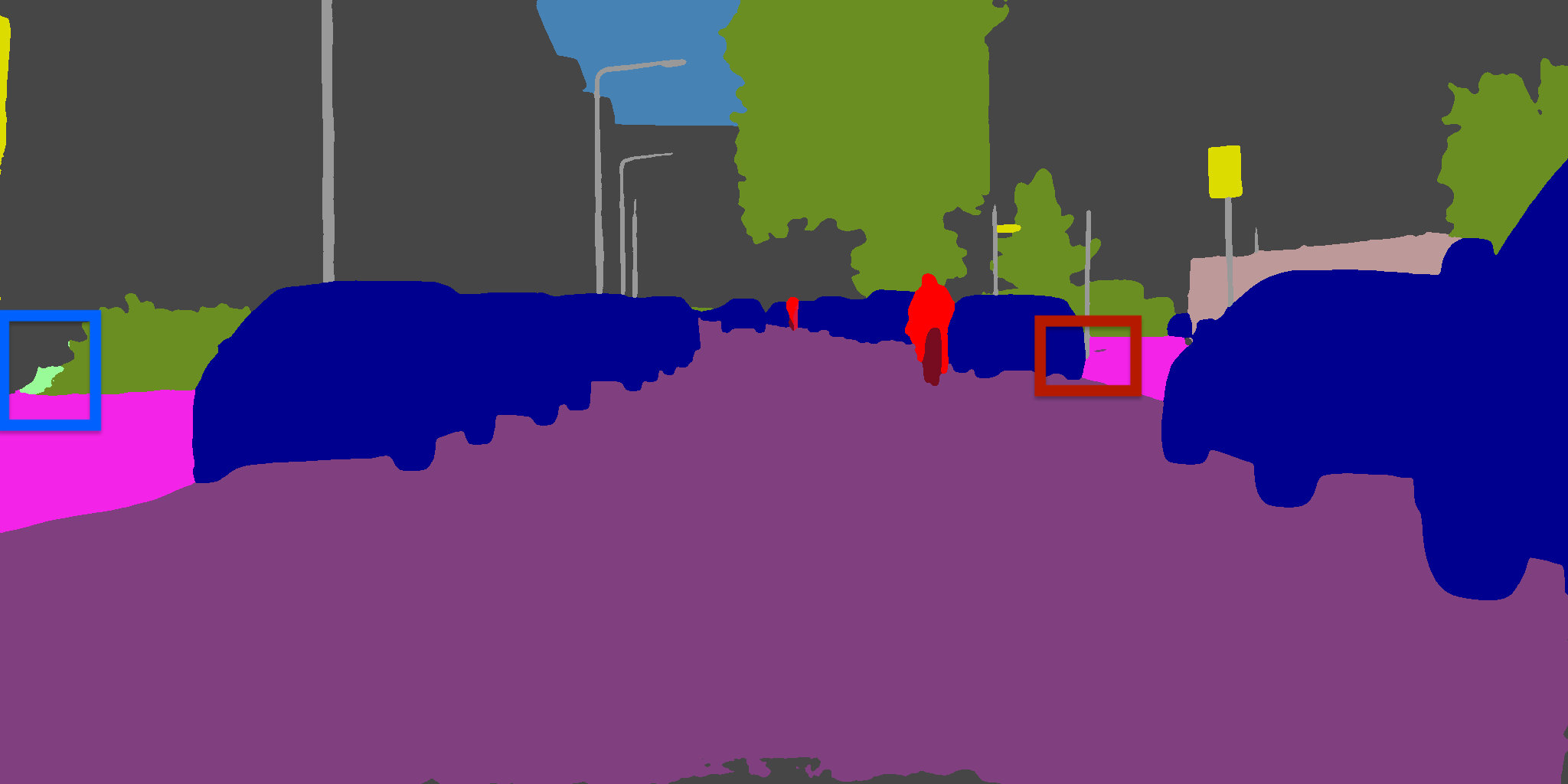}}
  \hfil
  \subfloat{\includegraphics[width=0.165\linewidth]{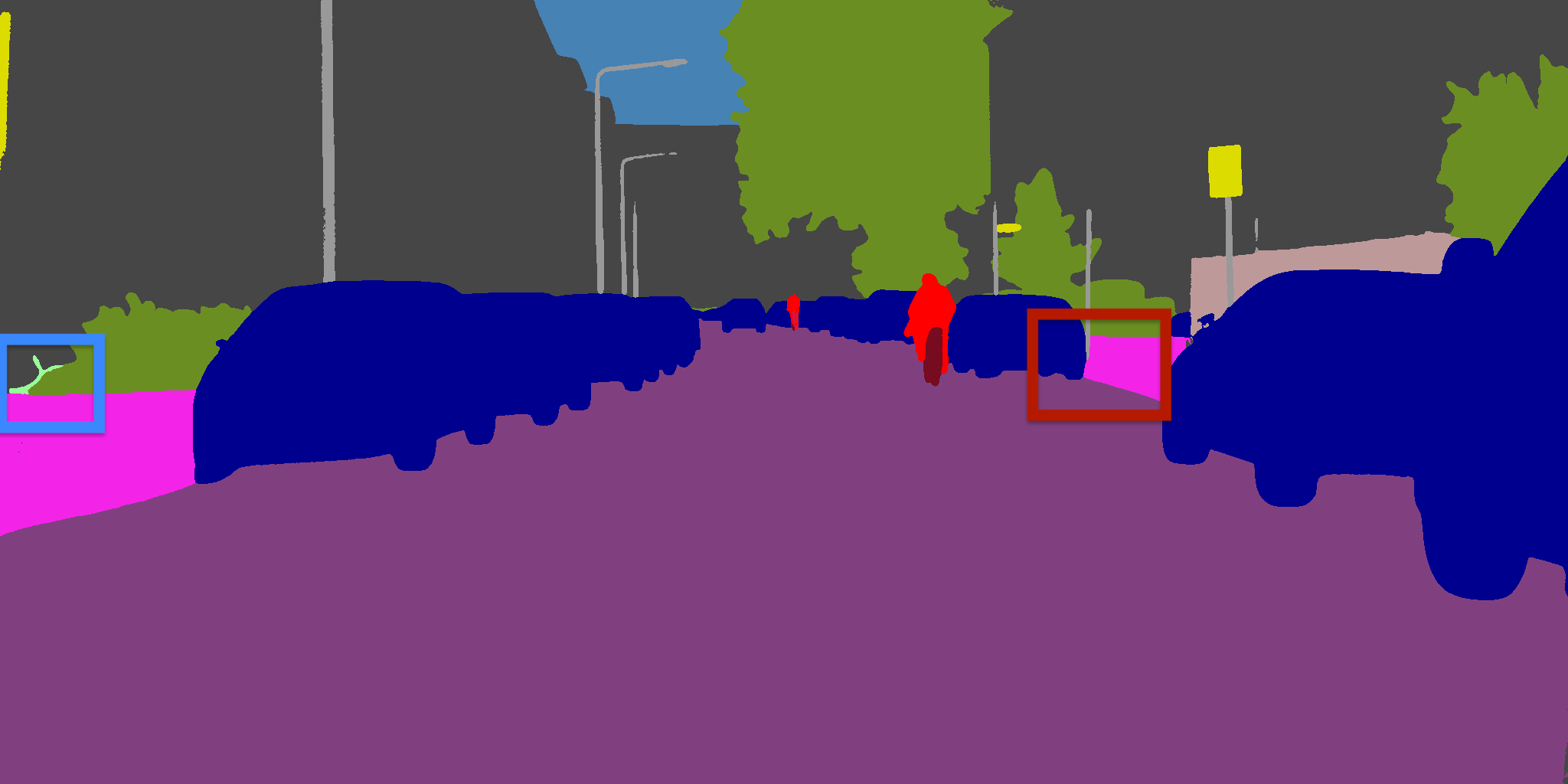}}
  \vspace{-0.35cm}
  \subfloat{\includegraphics[width=0.165\linewidth]{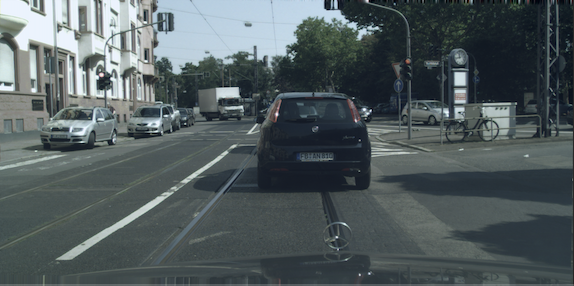}}
  \hfil
  \subfloat{\includegraphics[width=0.165\linewidth]{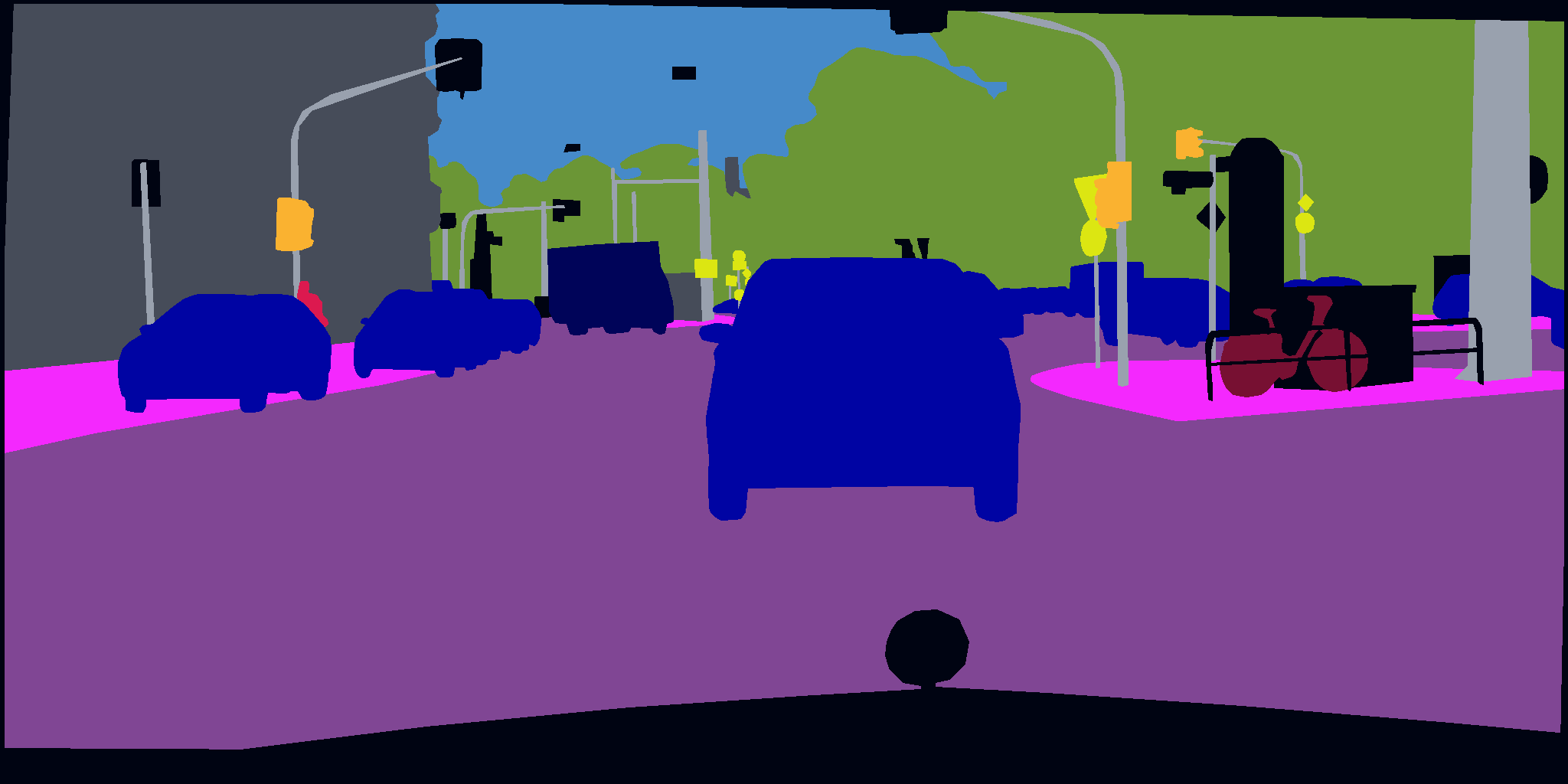}}
  \hfil
  \subfloat{\includegraphics[width=0.165\linewidth]{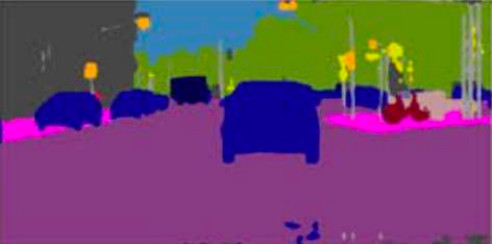}}
  \hfil
  \subfloat{\includegraphics[width=0.165\linewidth]{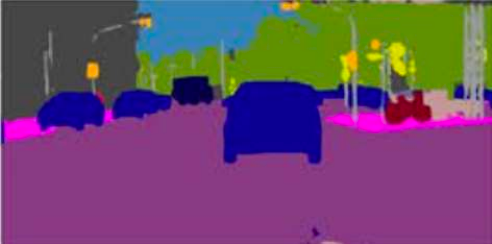}}
  \hfil
  \subfloat{\includegraphics[width=0.165\linewidth]{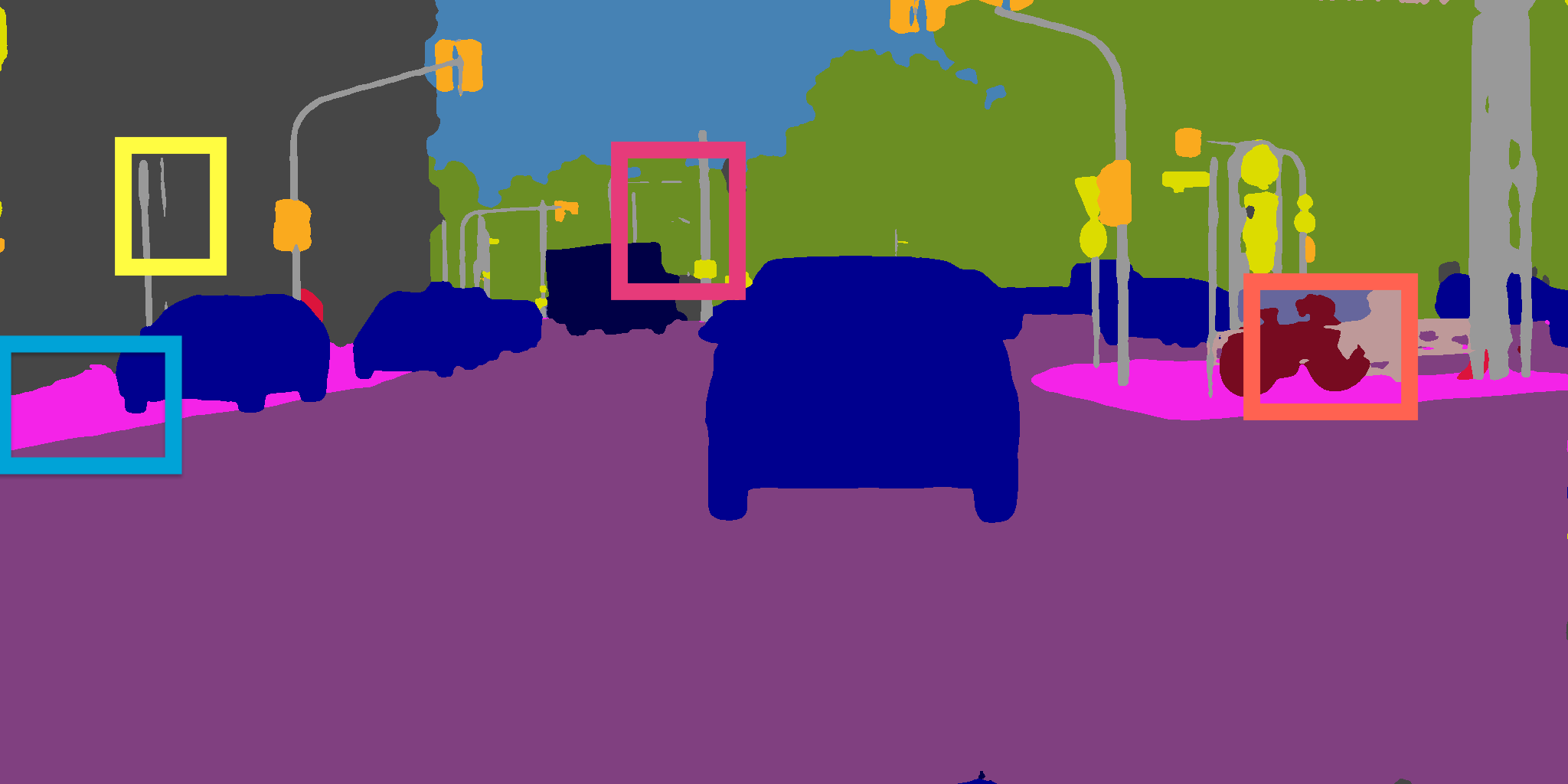}}
  \hfil
  \subfloat{\includegraphics[width=0.165\linewidth]{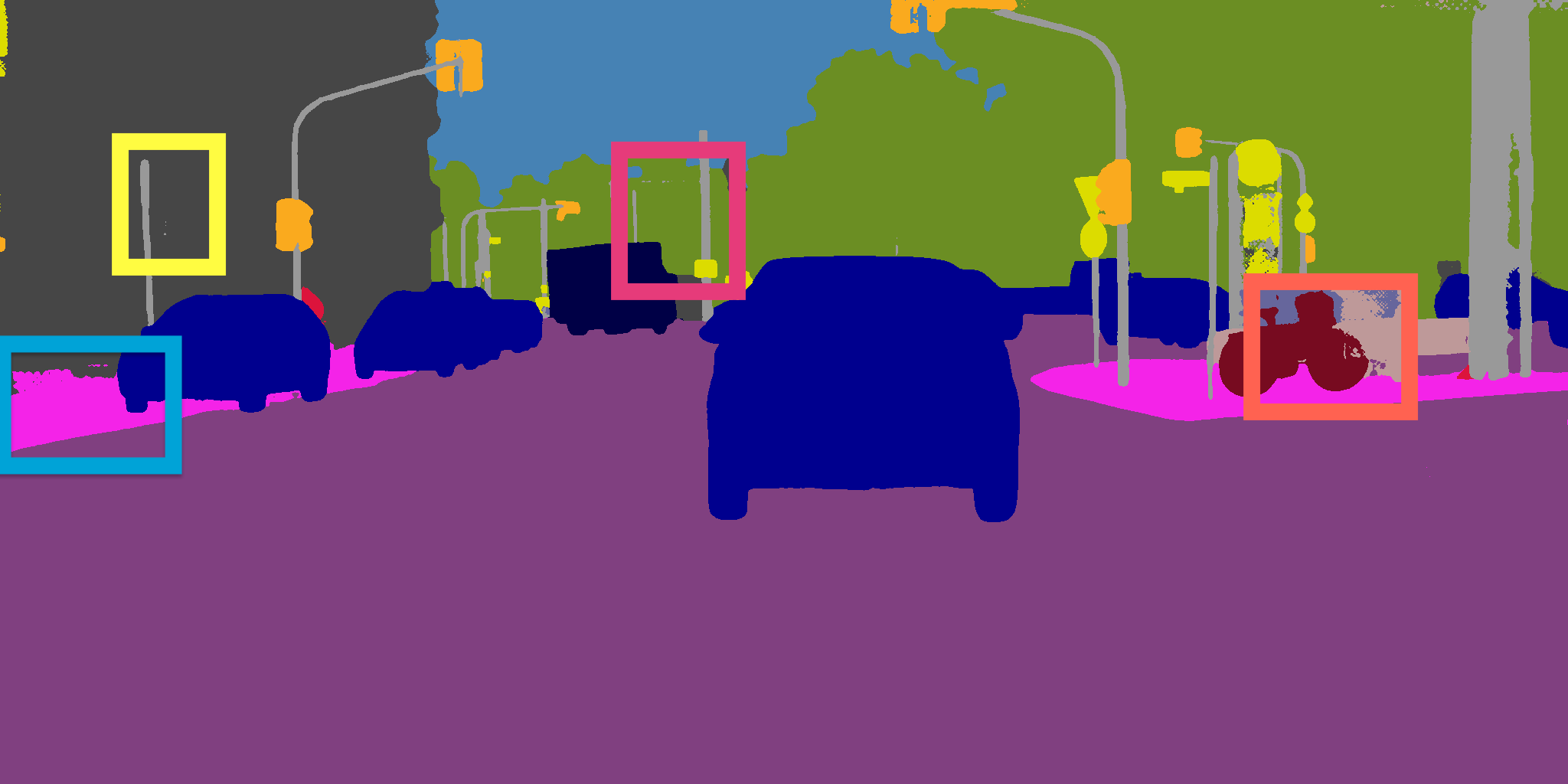}}
     \vspace{-0.35cm}
  \subfloat{\includegraphics[width=0.165\linewidth]{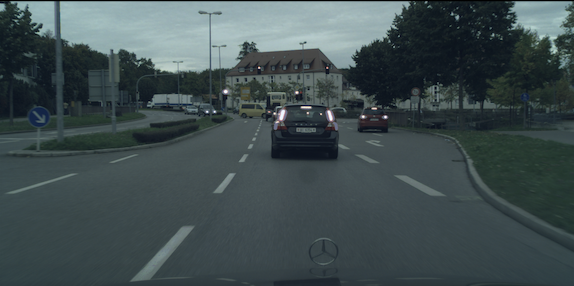}}
  \hfil
  \subfloat{\includegraphics[width=0.165\linewidth]{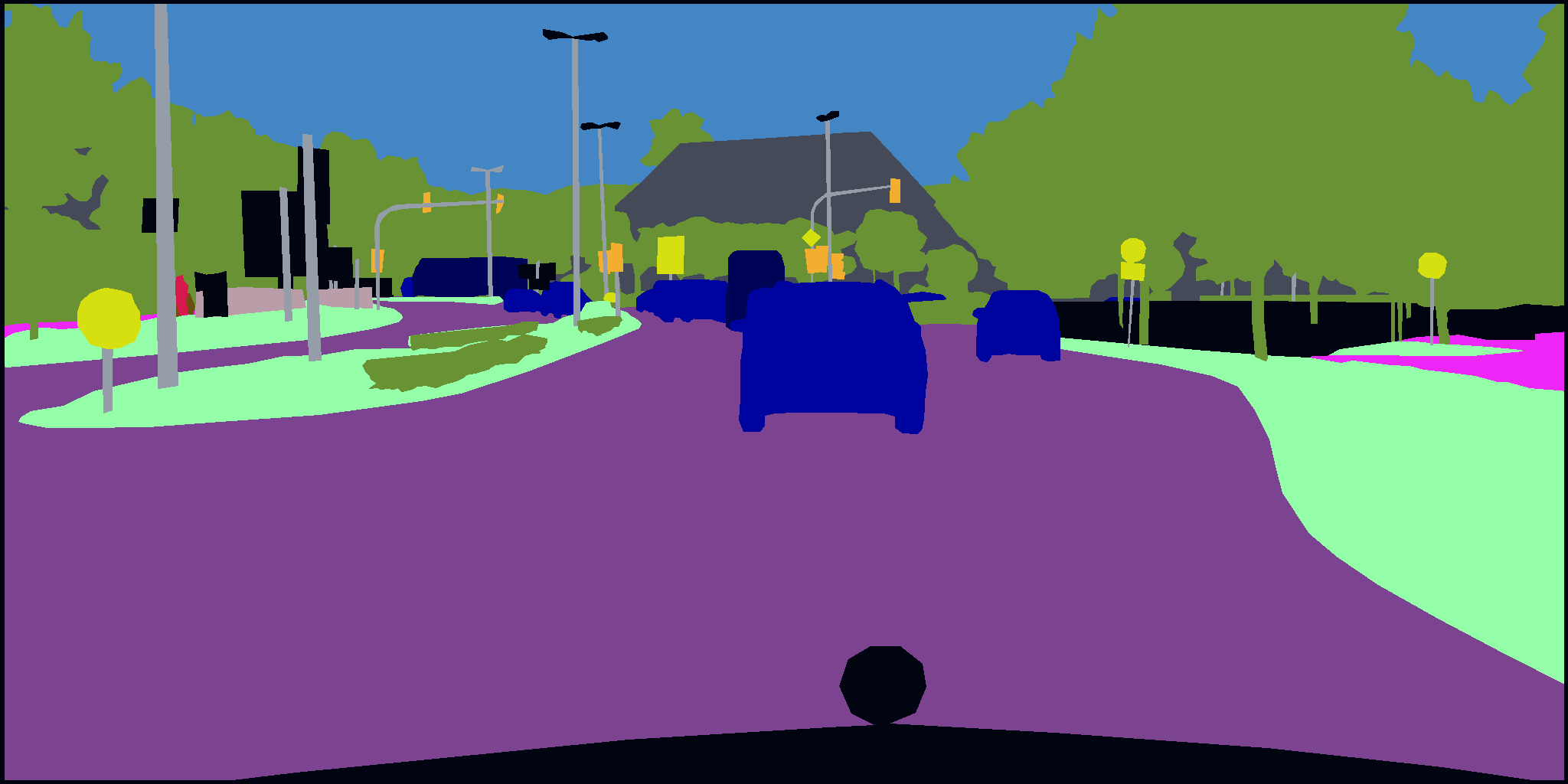}}
  \hfil
  \subfloat{\includegraphics[width=0.165\linewidth]{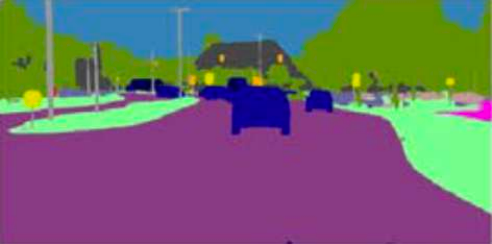}}
  \hfil
  \subfloat{\includegraphics[width=0.165\linewidth]{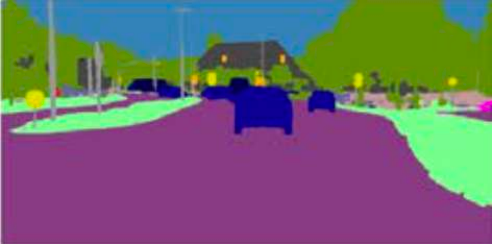}}
  \hfil
  \subfloat{\includegraphics[width=0.165\linewidth]{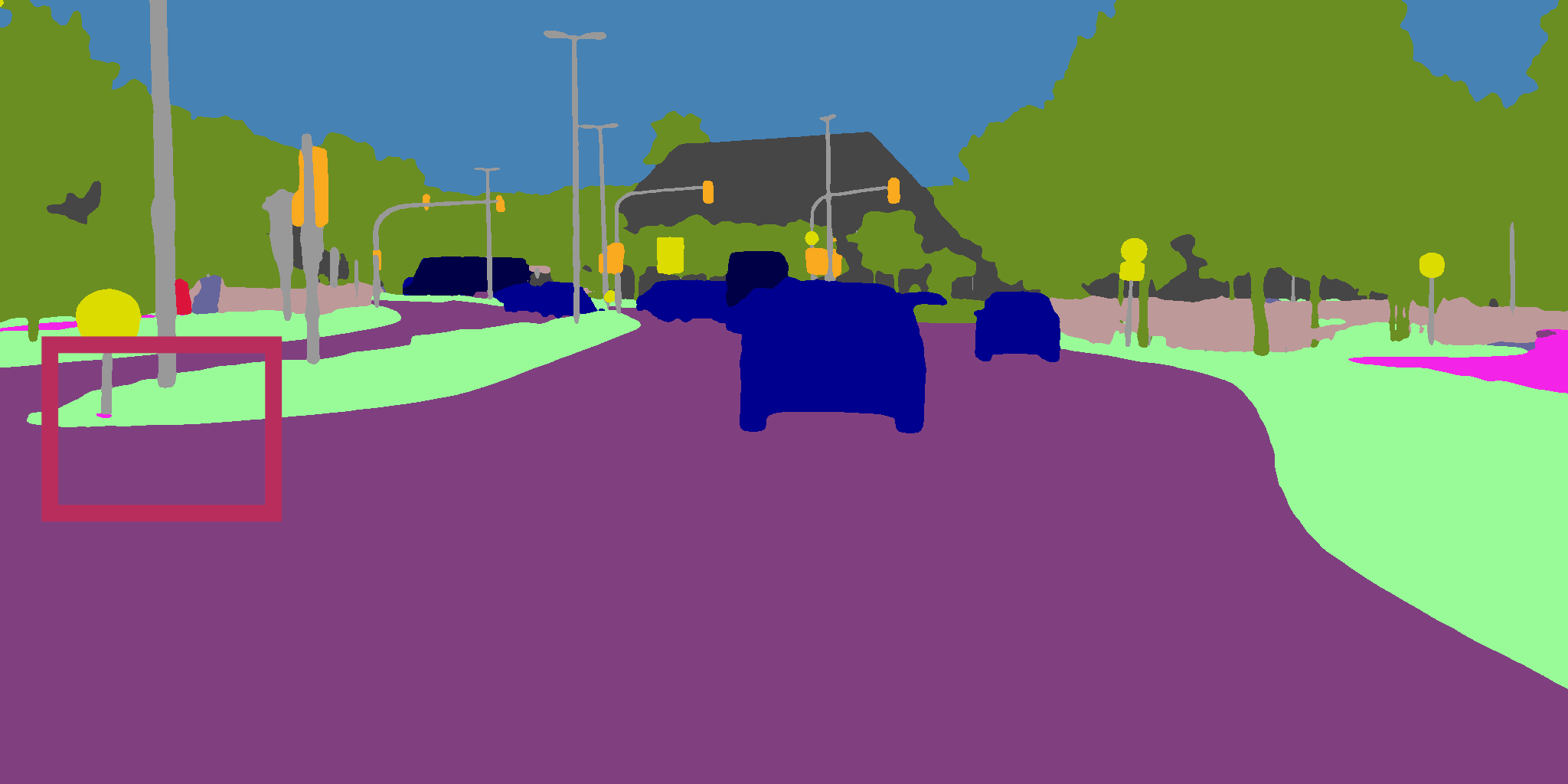}}
  \hfil
  \subfloat{\includegraphics[width=0.165\linewidth]{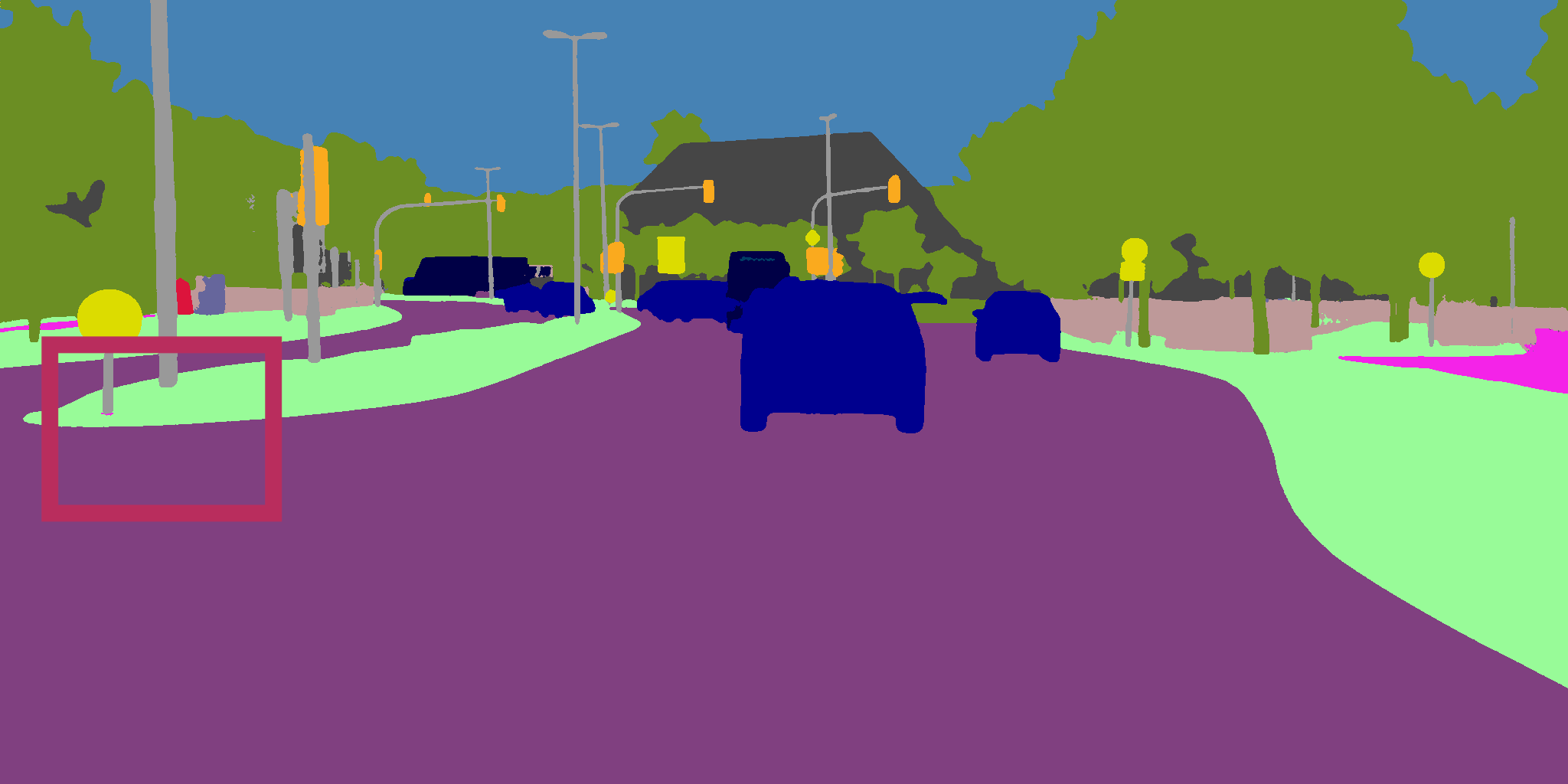}}
    \vspace{-0.35cm}
  \subfloat{\includegraphics[width=0.165\linewidth]{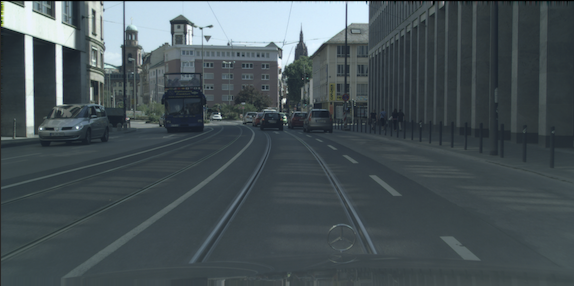}}
  \hfil
  \subfloat{\includegraphics[width=0.165\linewidth]{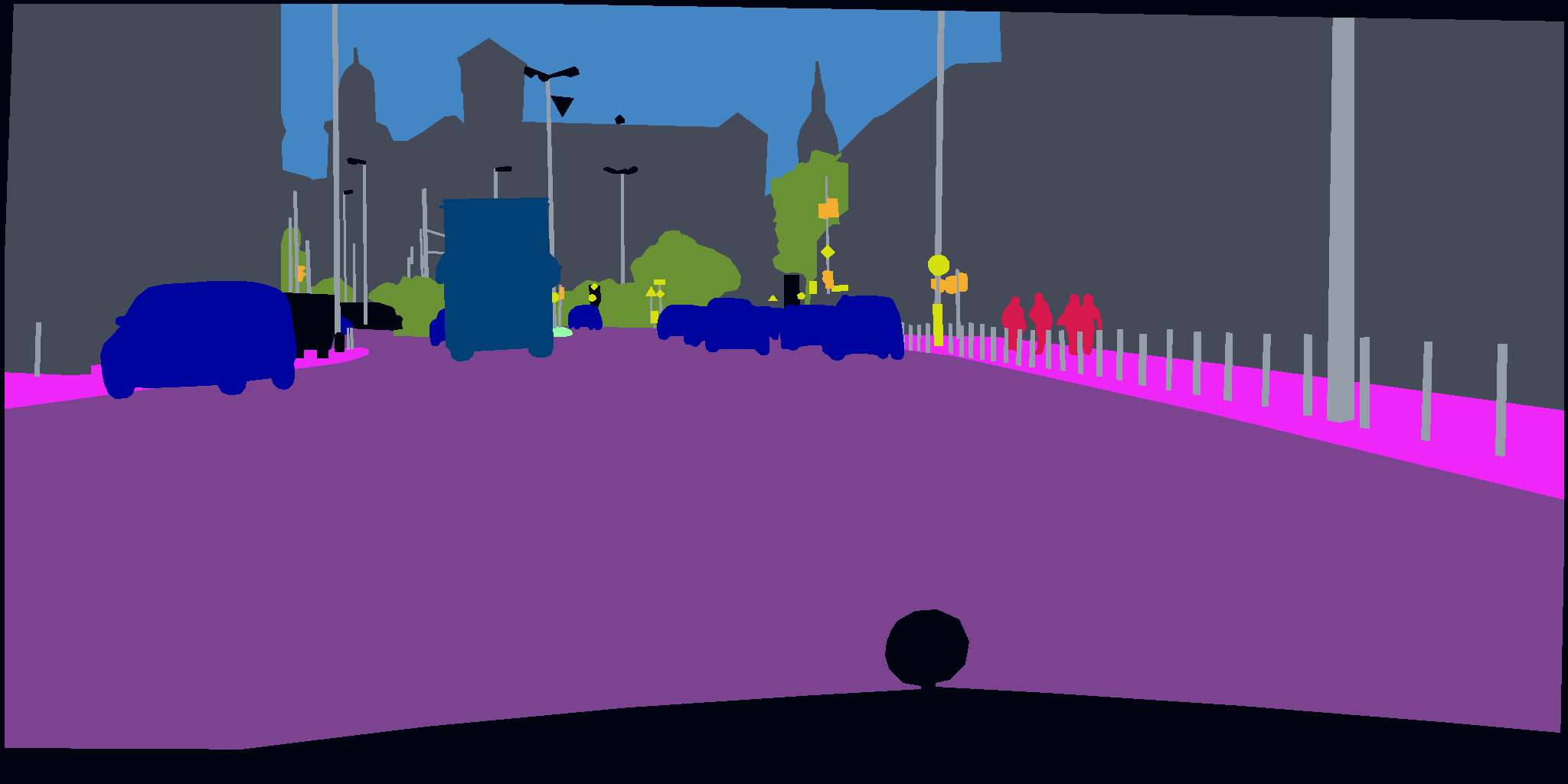}}
  \hfil
  \subfloat{\includegraphics[width=0.165\linewidth]{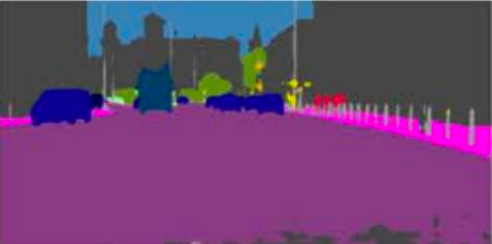}}
  \hfil
  \subfloat{\includegraphics[width=0.165\linewidth]{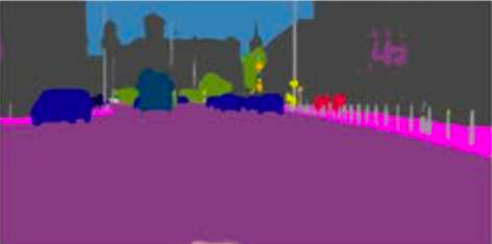}}
  \hfil
  \subfloat{\includegraphics[width=0.165\linewidth]{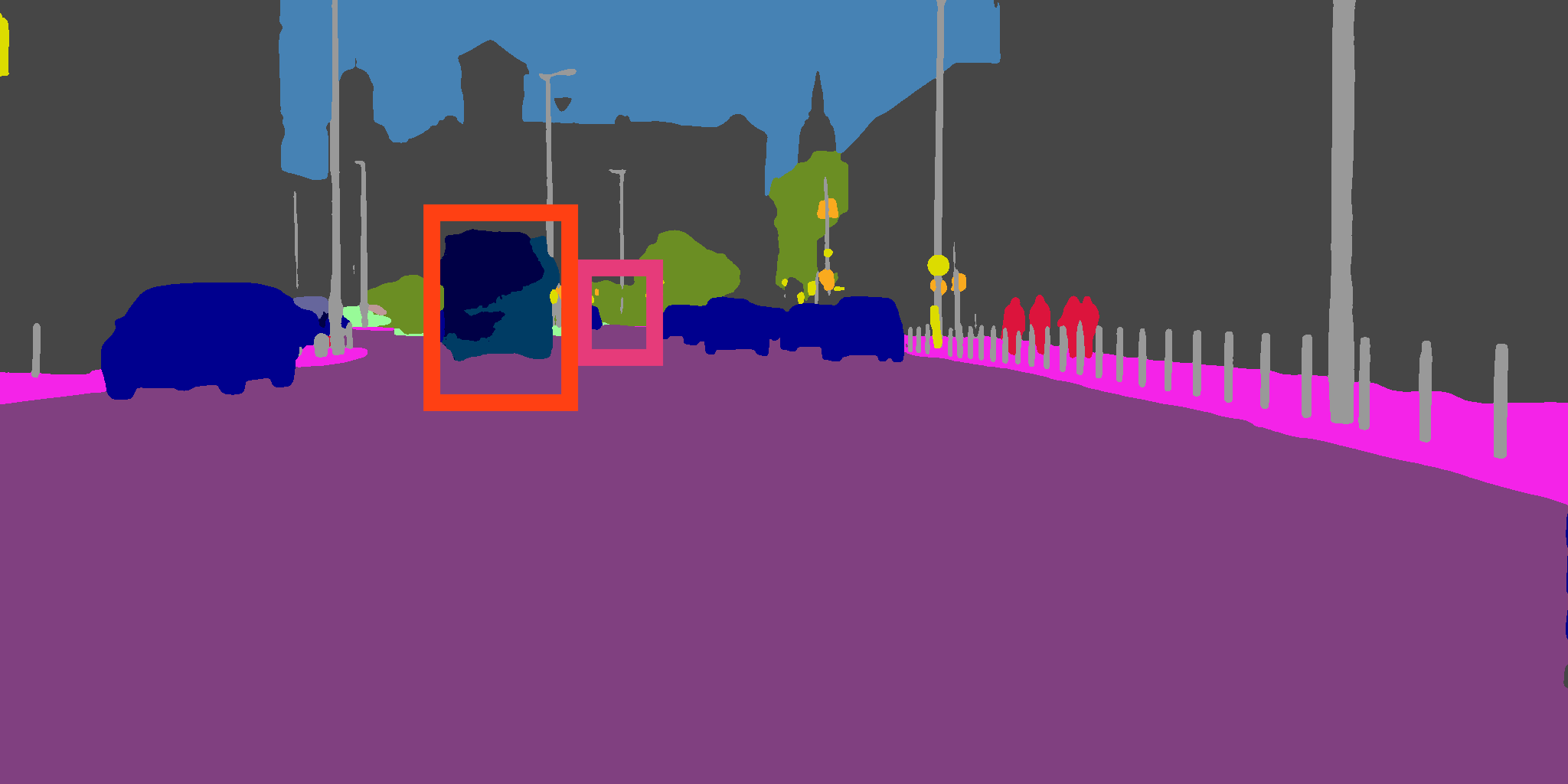}}
  \hfil
  \subfloat{\includegraphics[width=0.165\linewidth]{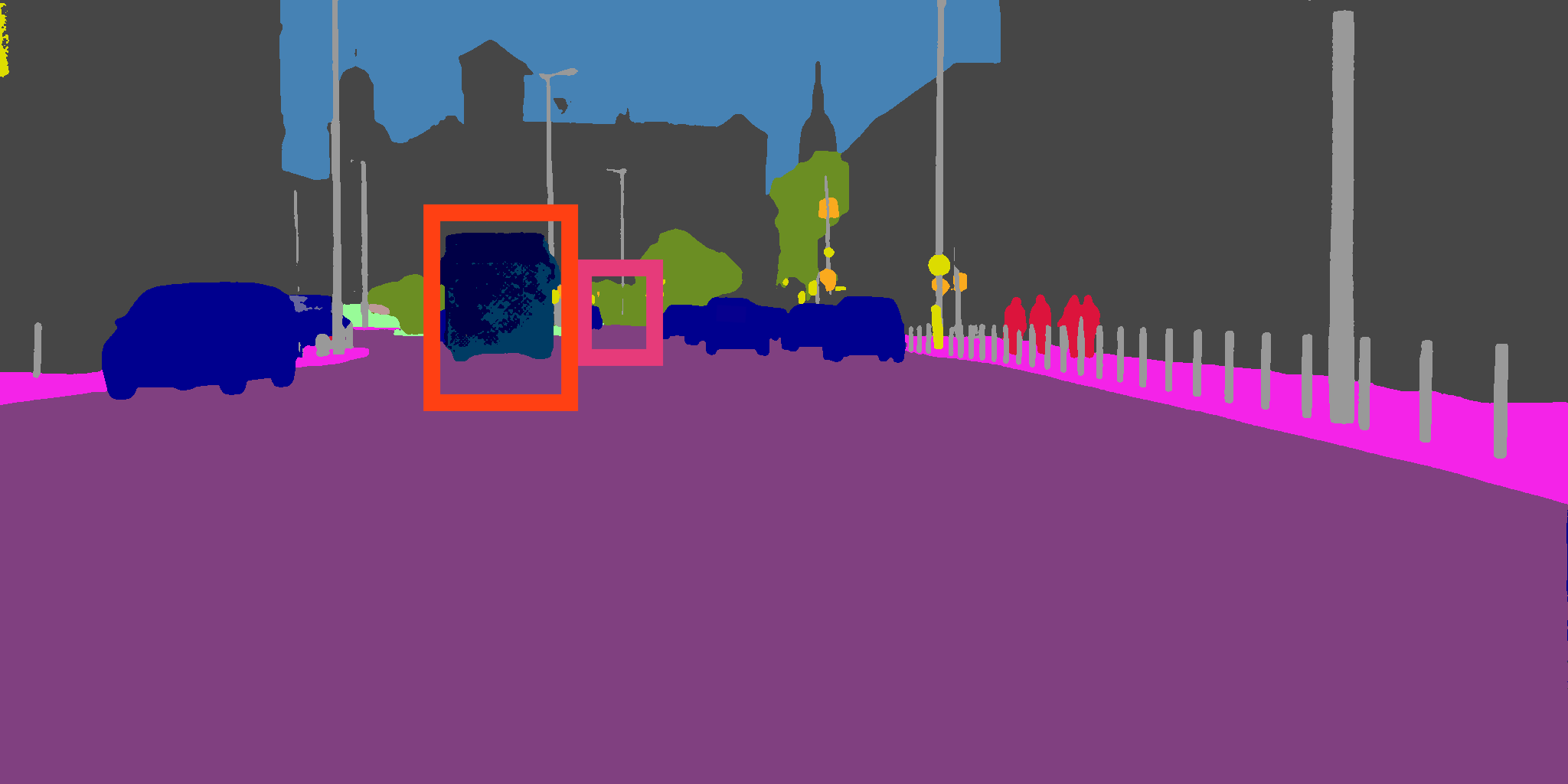}}
  \vspace*{1mm}
  \caption{ \textbf{Qualitative results of selected  examples on Cityscapes}.From left to right: RGB Input Image, GT, CCNet \cite{HuangWHHWL2019}, DANet \cite{FuLYL19}, HRNet \cite{wang2020deep}, OVeNet. Best viewed on a screen and zoomed in.}
  \label{fig:comp_city}
  \vspace*{-0.3cm}
\end{figure*}

Qualitative results on Cityscapes support the above findings, as shown in Fig.~\ref{fig:comp_city}. To be more specific, from left to right, we depict the RBG input image, GT image, CCNet's \cite{HuangWHHWL2019}, DANet's \cite{FuLYL19}, HRNet's \cite{YuanCW18} and our model's prediction. Specifically,  our model demonstrates successful classification of incorrectly predicted pixels (identified by a red and a blue frame) in the first example. In the second example, OVeNet exhibits superior performance compared to previous models, as it accurately expands the sidewalk and the pole depicted in the blue and yellow frames, respectively. Furthermore, not only does it successfully eliminate discontinuities in the pink frame, but it also predicts a better representation of the bicycle's shape. As far as the last two examples are concerned, our model shrinks the false predictions of HRNet on sidewalk, pole and bus segments resulting in a superior prediction compared to the HRNet baseline. To sum up, OVeNet surpasses the performance of HRNet.

\setlength{\tabcolsep}{3pt}
\begin{table}[t]
    \scriptsize
    \centering 
    \caption{\textbf{Semantic segmentation results on Cityscapes test set.}. We compare our method against SOTA methods as in \cite{wang2020deep}.D-ResNet-101 = Dilated-ResNet-101. By default, OVeNet is built on HRNet, unless stated otherwise.}
    \label{tab:test_cityscapes}
    \vspace*{1mm}
    \resizebox{\linewidth}{!}{%
    \begin{tabular}{l|l|cccc}
        \toprule%
          & backbone & mIoU  & iIoU cla. & IoU cat. & iIoU cat.\\
        \midrule
  
        \multicolumn{3}{l}{\emph {Model learned on the \texttt{train} set}}\\
        \midrule
        PSPNet~\cite{ZhaoSQWJ17} & D-ResNet-$101$ & $78.4$ & $56.7$ & $90.6$  & $78.6$ \\
        PSANet~\cite{ZhaoZLSLLJ18} & D-ResNet-$101$ & $78.6$ & - & - & - \\
        HRNet \cite{wang2020deep}  & HRNetV$2$-W$48$ & ${80.4}$ & ${59.2}$ & ${91.5}$ & ${80.8}$\\
        \midrule
        OVeNet & HRNetV2-W48 & $\mathbf{81.8}$ &$\mathbf{61.6}$ & $\mathbf{91.9}$ &$\mathbf{82.2}$ \\
      
        \midrule
        \multicolumn{3}{l}{
        \emph {Model learned on the \texttt{train+val} set}}\\
        \midrule
        DeepLab~\cite{ChenPKMY18} & D-ResNet-$101$ & $70.4$ & $42.6$ & $86.4$ & $67.7$\\
        RefineNet~\cite{LinMSR17}& ResNet-$101$ & $73.6$ & $47.2$ & $87.9$ & $70.6$\\
        DSSPN~\cite{LiangZX18} & D-ResNet-$101$ & $76.6$ & $56.2$ & $89.6$ & $77.8$\\
        ResNet38~\cite{WuSH16e} & WResNet-38 &$78.4$ &$59.1$ &$90.9$ &$78.1$ \\
        PADNet~\cite{OWS18} & D-ResNet-$101$ & $80.3$ & $58.8$ & $90.8$ & $78.5$\\
        CFNet~\cite{ZhangZWX} & D-ResNet-$101$ & $79.6$ & - & - & -\\
        Auto-DeepLab~\cite{liu2019auto} & - & $80.4$ & - & - & -\\
        DenseASPP~\cite{ZhaoSQWJ17} & WDenseNet-$161$ & $80.6$ & $59.1$ & $90.9$ & $78.1$ \\
        CCNet~\cite{HuangWHHWL2019} & D-ResNet-$101$ & $81.4$ & - & - & -\\
        DANet~\cite{FuLYL19} & D-ResNet-$101$ & $81.5$ & - & - & - \\
        HRNet \cite{wang2020deep}  &  HRNetV$2$-W$48$  & $81.6$ & $61.8$ & $92.1$ & $82.2$ \\
        HRNet + OCR~\cite{YuanCW18} & HRNetV$2$-W$48$ &  $81.9$ & $\mathbf{62.0}$ &$\mathbf{92.0}$ & $\mathbf{82.5}$\\
        \midrule
        OVeNet (\emph{HRNet $+$ OCR})  & HRNetV2-W48 & $\mathbf{82.4}$ & 61.6 & 91.9 &$82.2$ \\
        \bottomrule
    \end{tabular}
    }
    \vspace{-0.4cm}
\end{table}

\newcommand*\rot{\rotatebox{90}}

\begin{table*}[htpb!]
    \centering
    \caption{\textbf{Per Class Results on Cityscapes test set}}
    \label{tab:class}

    \resizebox{\linewidth}{!}{%
    \begin{tabular}{lc*{19}c}
        \toprule
        Method & \rot{road} & \rot{sidew.} & \rot{build.} & \rot{wall} 
        & \rot{fence} & \rot{pole} & \rot{light} 
        & \rot{sign} & \rot{veget.} & \rot{terrain} & \rot{sky} & \rot{person} & \rot{rider} & \rot{car} & \rot{truck}& \rot{bus}& \rot{train}& \rot{motorc.} & \rot{bicycle} & mIoU \\
        \midrule
        HRNet \cite{wang2020deep}  & $98.73$ & $\mathbf{87.49}$ & $93.65$ & $56.48$ & $61.57$ & $\mathbf{71.57}$ & $78.76$ & $81.81$ & $93.99$ & $74.11$ & $95.68$ & $87.95$ & $73.72$ & $96.35$ & $69.94$ & $82.52$ & $76.93$ & $70.88$ & $\mathbf{78.02}$ & $80.40$\\
        
        OVeNet & $\mathbf{98.74}$ & $87.41$ & $\mathbf{93.79}$ & $\mathbf{61.65}$ & $\mathbf{64.00}$ & $71.35$ & $\mathbf{78.98}$ & $\mathbf{81.65}$ & $\mathbf{94.00}$ & $\mathbf{73.42}$ & $\mathbf{95.81}$ & $\mathbf{87.99}$ & $\mathbf{74.36}$ & $\mathbf{96.42}$ & $\mathbf{74.76}$ & $\mathbf{87.70}$ & $\mathbf{82.83}$ & $\mathbf{71.77}$ & $77.86$ & $\mathbf{81.82}$\\
        \midrule
        HRNet $+$ OCR & $98.77$ & $\mathbf{87.85}$ & $93.72$ & $57.75$ & $63.92$ & $\mathbf{71.74}$ & $\mathbf{78.56}$ & $\mathbf{81.77}$ & $\mathbf{94.06}$ & $\mathbf{73.69}$ & $95.68$ & $88.04$ & $74.64$ & $96.46$ & $76.40$ & $88.78$ & $84.63$ & $71.79$ & $\mathbf{78.63}$ & $81.90$\\

        OVeNet (\emph{HRNet $+$ OCR})  & $\mathbf{98.79}$ & $87.47 $ &$ \mathbf{93.86} $ &$\mathbf{ 62.97} $ &$\mathbf{ 64.41 }$ &$ 70.80 $ &$ 78.45 $ &$ 81.13 $ &$ 93.99 $ &$ 73.31 $ &$ \mathbf{95.72} $ &$ \mathbf{88.08} $ &$ \mathbf{74.90} $ &$ \mathbf{96.47} $ &$\mathbf{ 76.95} $ &$\mathbf{ 89.95} $ &$ \mathbf{88.48} $ &$ \mathbf{71.87} $ &$ 78.43$& $\mathbf{82.42}$\\
        \bottomrule
    \end{tabular}
    }
    \vspace{-0.3cm}
\end{table*}

\PAR{ACDC.} We also outperform initial HRNet and prior SOTA methods under similar training time. Table \ref{tab:conditions}  compares our approach with SOTA models not only on all methods but also on different conditions of the ACDC test set. OVeNet improves the initial HRNet model by $2.5\%$ mIoU in "All" conditions. Additionally, we can observe from the per class results shown on Table \ref{tab:class_acdc} on different conditions, that our approach outperforms HRNet in the vast majority of classes as well as in the total mIoU score. Specifically, in foggy images, small-instance classes like person, rider and  bicycle perform poorly  because of contrast reduction and resolution issues due to distant instances. There is also a huge performance boost of approximately $10\%$ and $15\%$ on the "bus" and "truck" class respectively. Moreover, it is more difficult to separate classes at night that are often dark or poorly lit, such as buildings, vegetation, traffic signs, and the sky. This behaviour is observed also in offset vector performance as they have small values when the visibility is limited. Lastly, during night and snow conditions, road and sidewalk performance is at its lowest, which can be attributed to misunderstanding between the two classes as a result of their similar look. As for val set results, HRNet achieves $75.5\%$ mIoU while our OVeNet surpasses it reaching $75.9\%$ mIoU.

\begin{figure}
  \centering
  \subfloat{\includegraphics[width=0.333\linewidth]{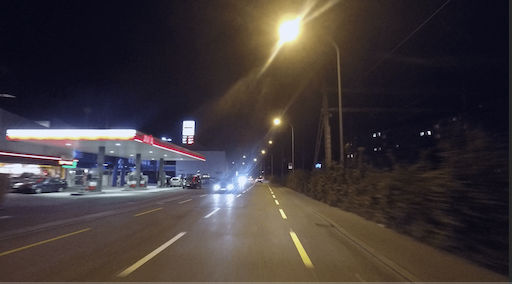}}
  \hfil
  \subfloat{\includegraphics[width=0.333\linewidth]{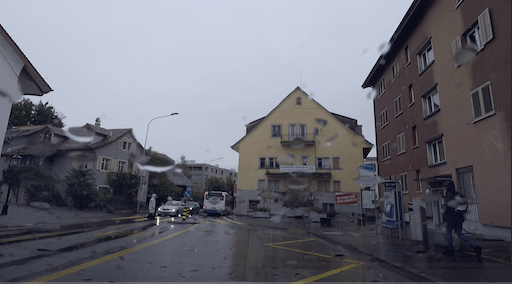}}
  \hfil
  \subfloat{\includegraphics[width=0.333\linewidth]{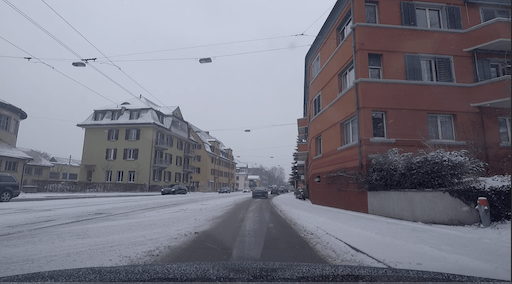}}

  \subfloat{\includegraphics[width=0.333\linewidth]{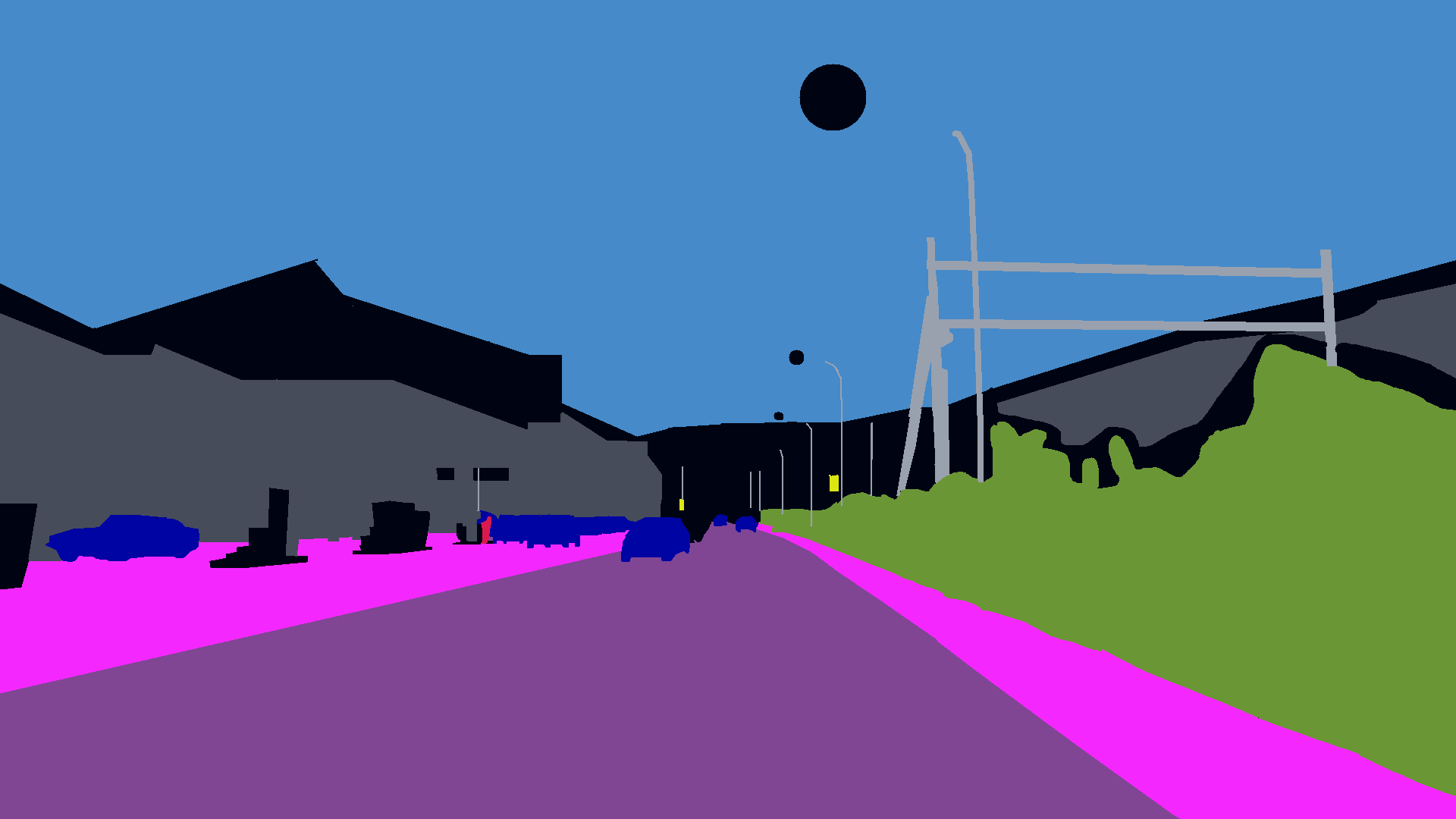}}
  \hfil
  \subfloat{\includegraphics[width=0.333\linewidth]{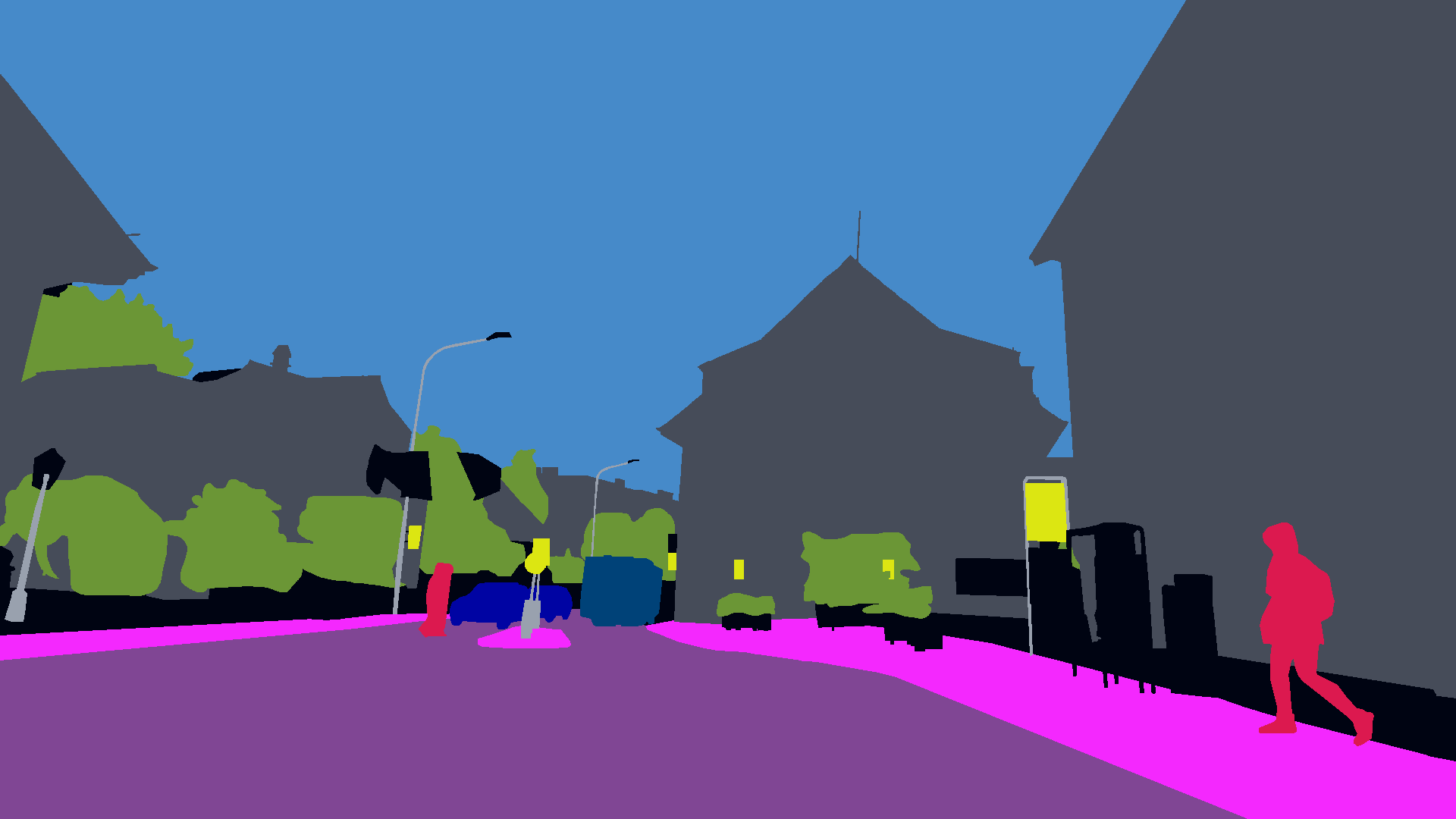}}
  \hfil
  \subfloat{\includegraphics[width=0.333\linewidth]{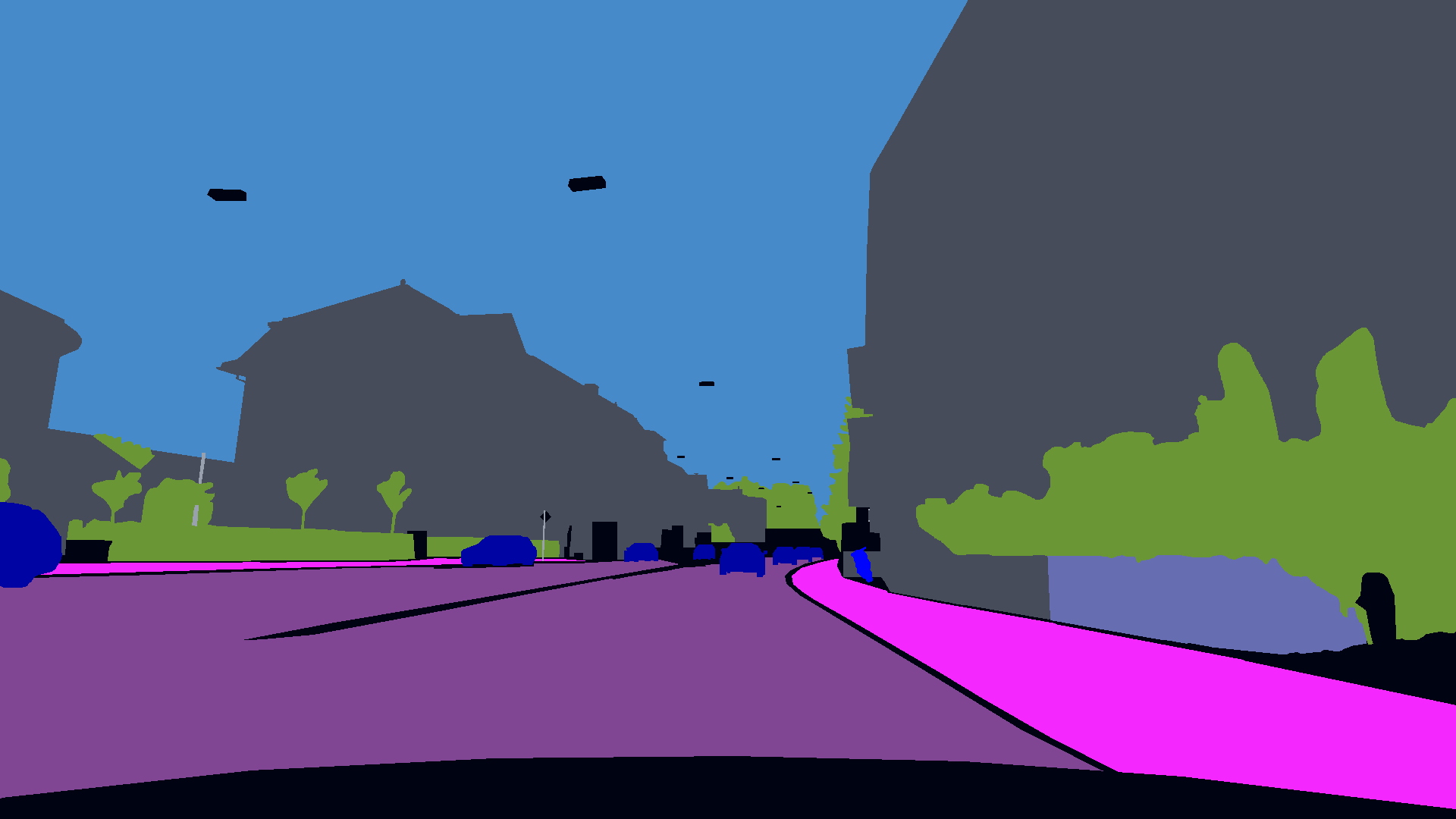}}

  \subfloat{\includegraphics[width=0.333\linewidth]{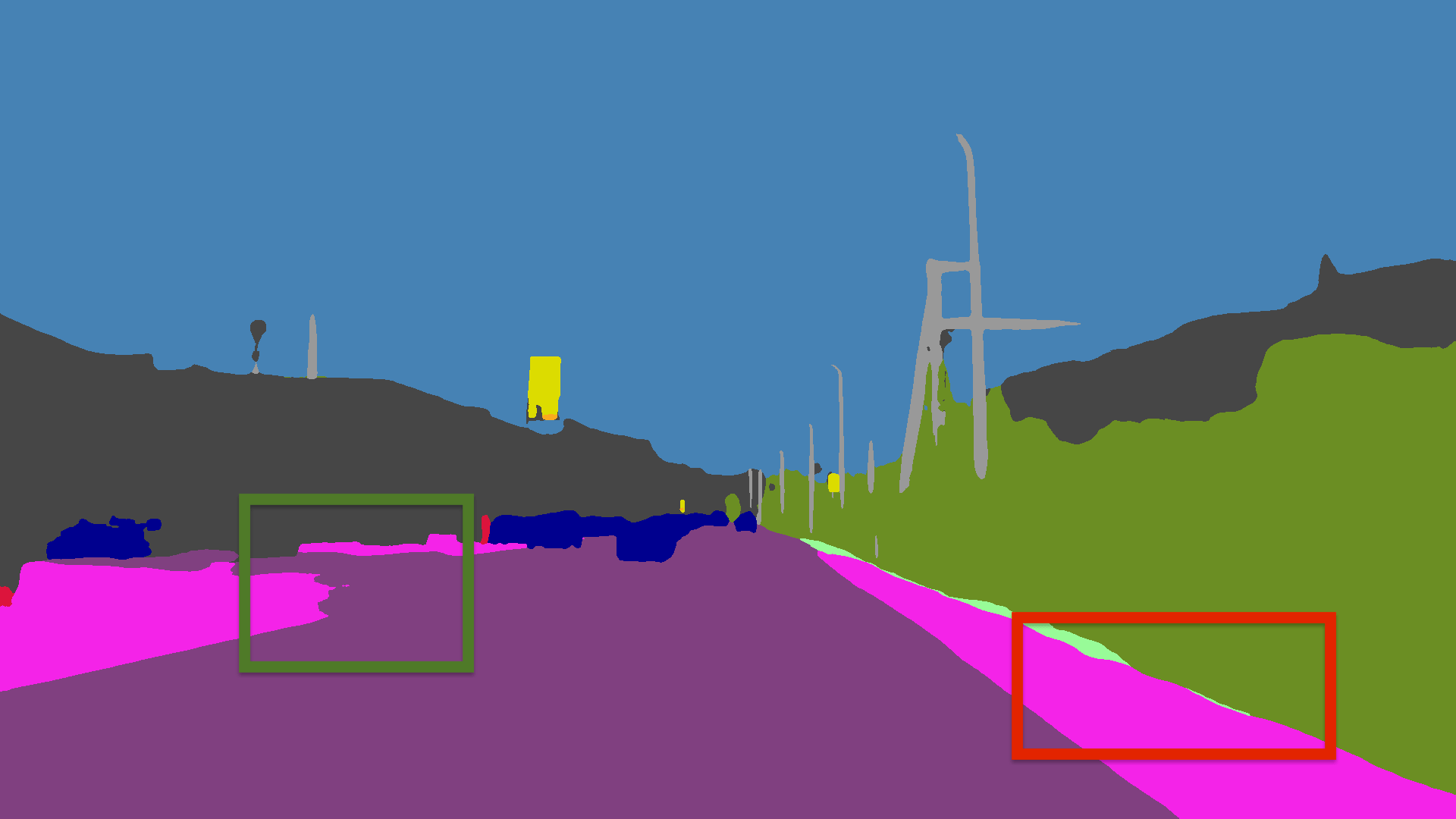}}
  \hfil
  \subfloat{\includegraphics[width=0.333\linewidth]{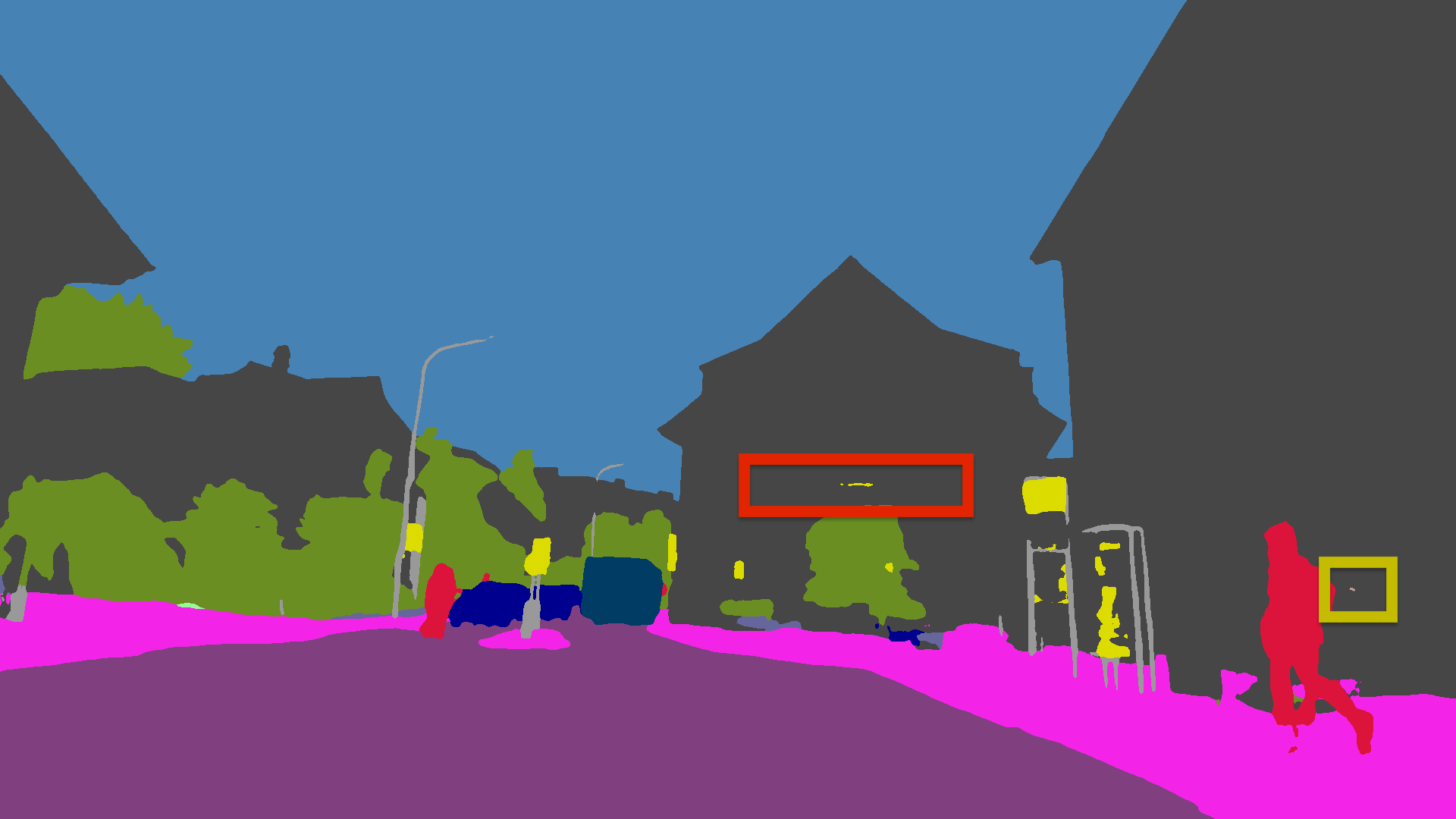}}
  \hfil
  \subfloat{\includegraphics[width=0.333\linewidth]{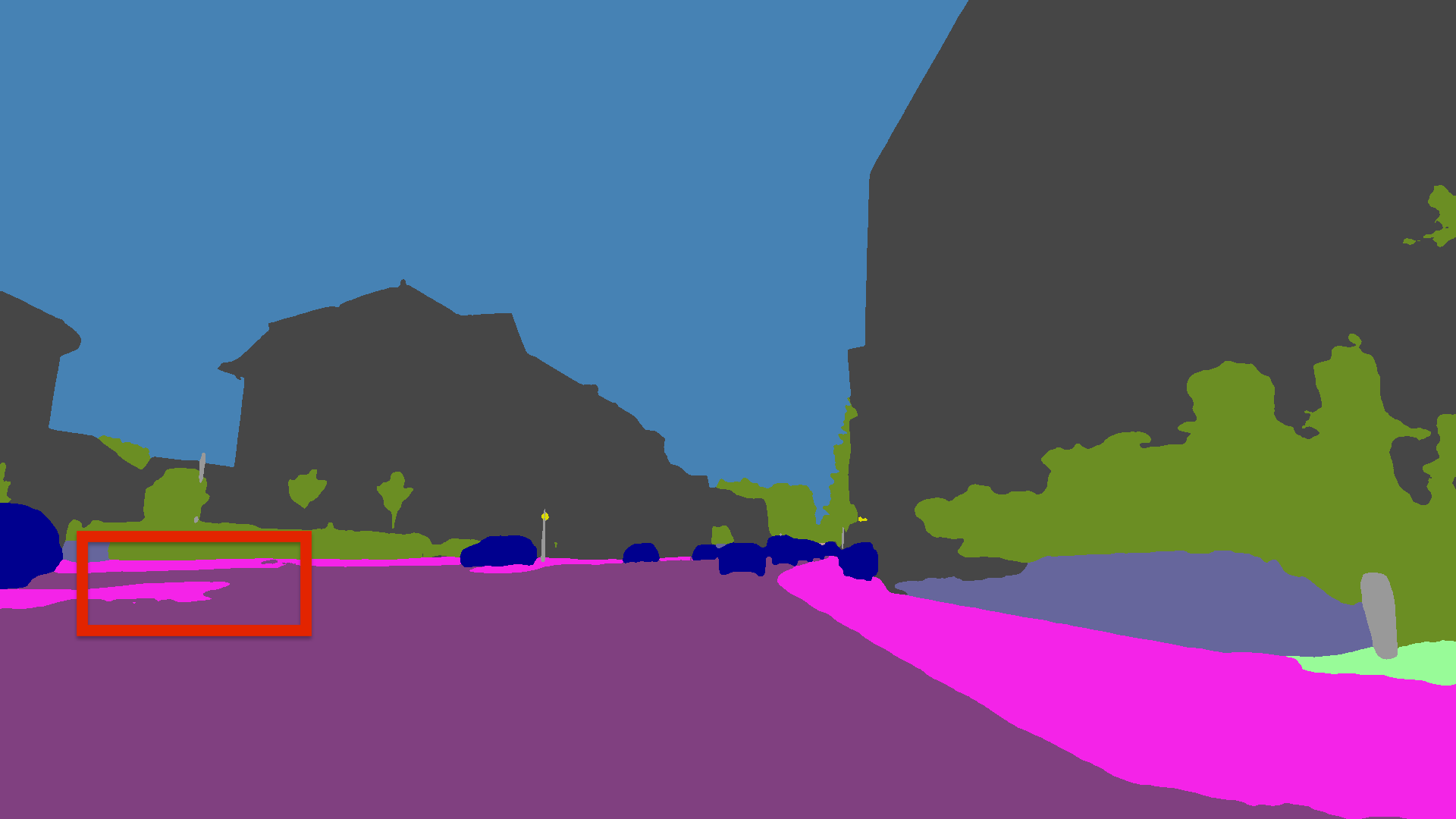}}

  \subfloat{\includegraphics[width=0.333\linewidth]{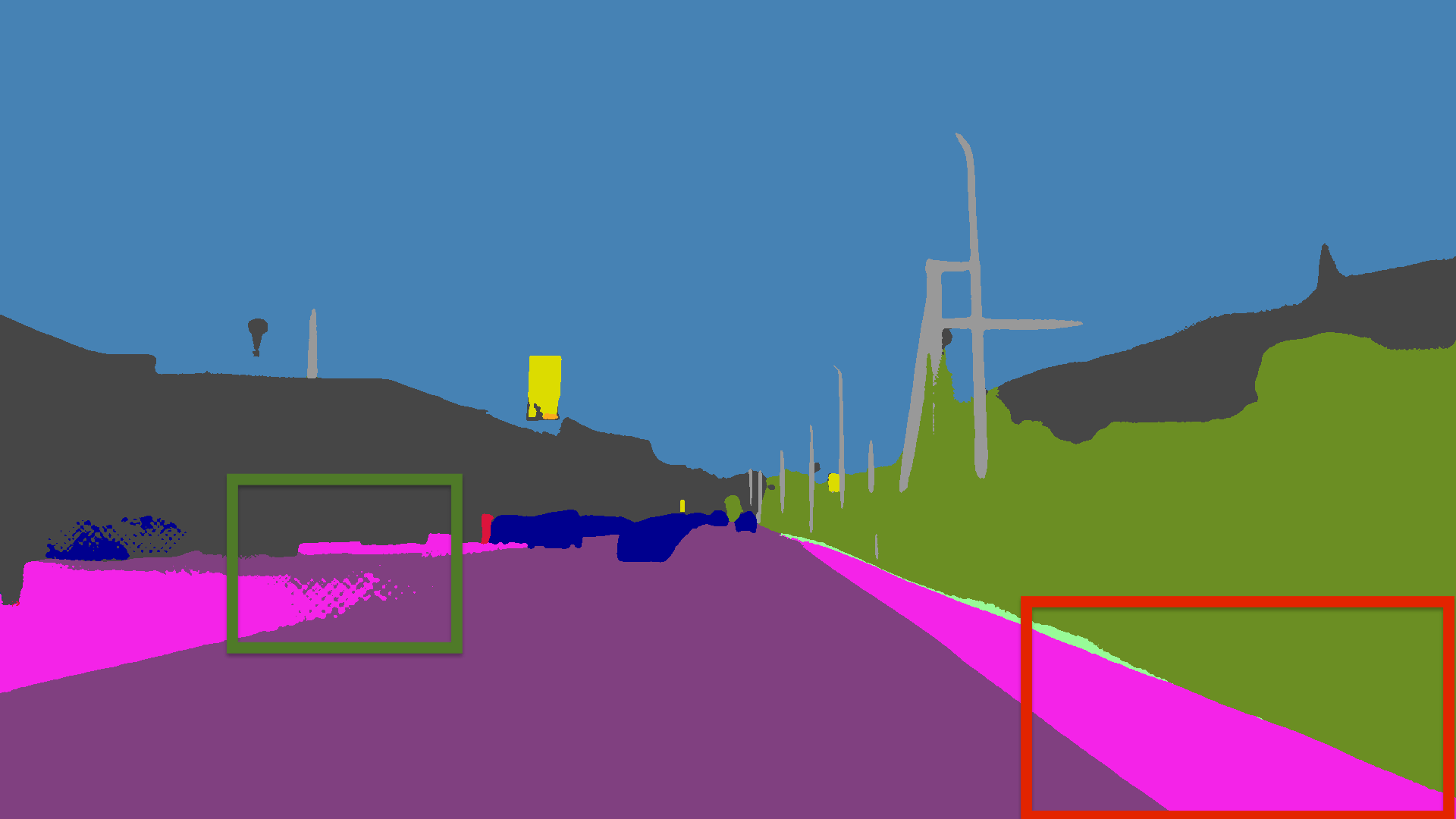}}
  \hfil
  \subfloat{\includegraphics[width=0.333\linewidth]{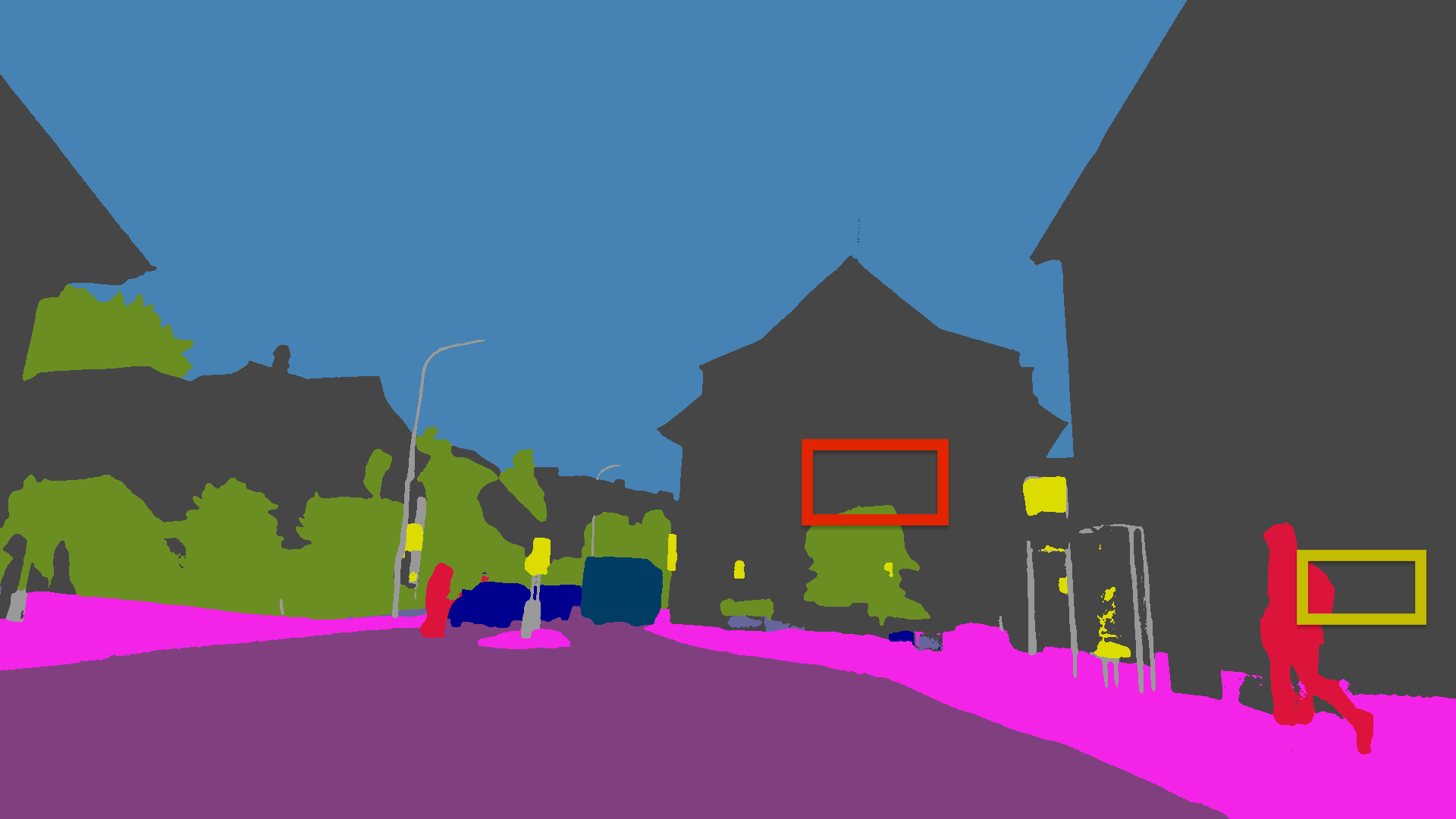}}
  \hfil
  \subfloat{\includegraphics[width=0.333\linewidth]{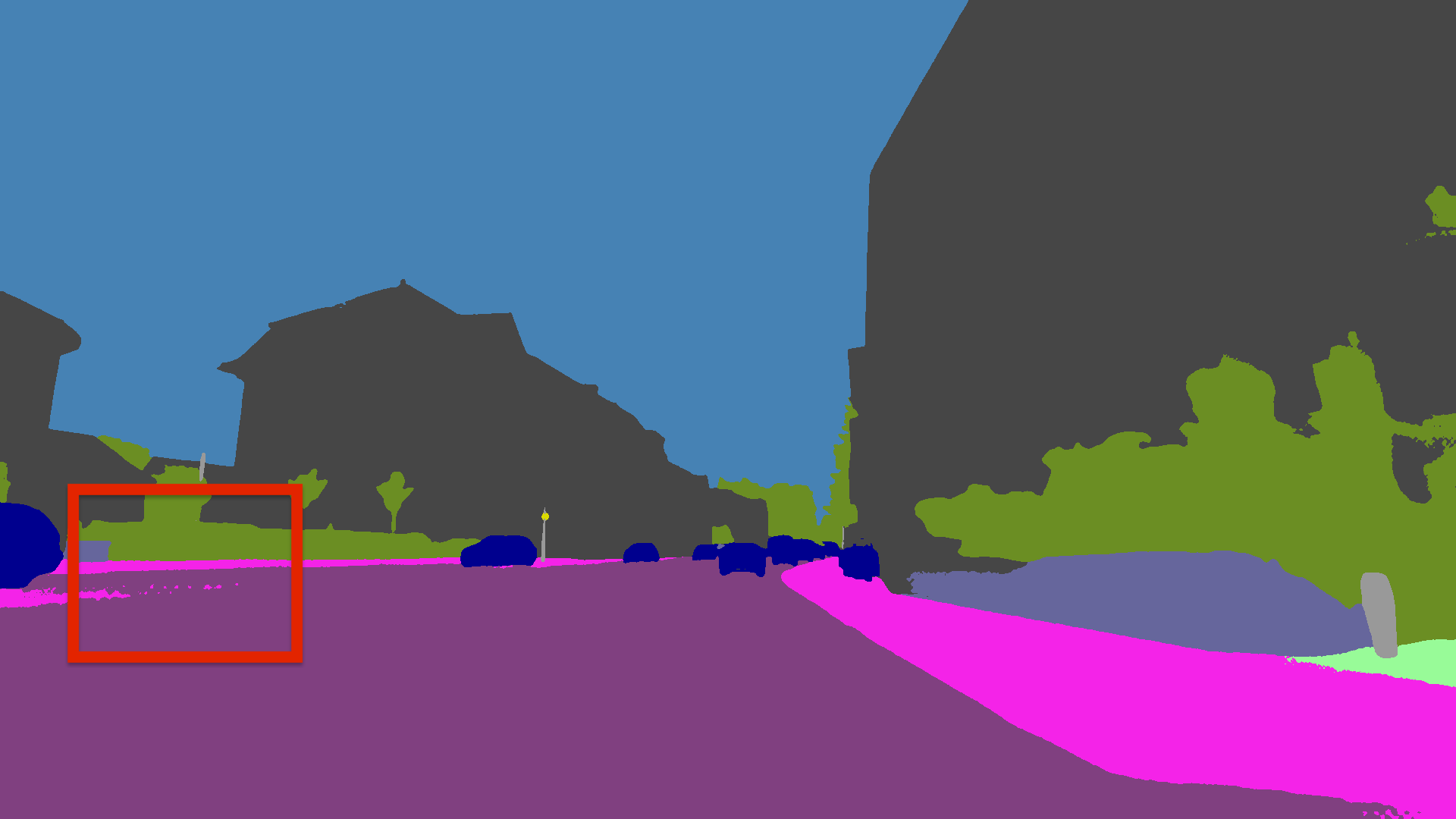}}
  \vspace*{1mm}
  \caption{ \textbf{Qualitative results of selected  examples on ACDC}. From top to bottom: RGB, GT, HRNet \cite{wang2020deep}, OVeNet. Best viewed on a screen and zoomed in.}
  \label{fig:classes}
  \vspace{-0.4cm}
\end{figure}

 Qualitative results on ACDC support the above findings, as shown in Fig. \ref{fig:classes}. To be more specific, in the first column, we can underline that our model tries to enlarge correctly the sidewalk segments in both red and green frames and reduces the erroneous terrain segment predicted by HRNet. As far as the second example is concerned, HRNet classifies incorrectly the sign of the house as traffic sign (red frame). On the contrary, our model corrects not only this mistake, but also a discontinuity occurring in the yellow frame. Last but not least, regarding the last set of materials, our offset vector-based model eliminates correctly the sidewalk area (red frame), which does not exist in the ground truth.

\begin{table*}
    \centering
    \caption{\textbf{Per Class Results on ACDC test set.} OVeNet is built on HRNet.}
    \label{tab:class_acdc}
      \vspace{-2.5mm}
    
    \resizebox{\linewidth}{!}{%
    \begin{tabular}{c|lc*{19}c}
        \toprule
        Condition &Method & \rot{road} & \rot{sidew.} & \rot{build.} & \rot{wall} 
        & \rot{fence} & \rot{pole} & \rot{light} 
        & \rot{sign} & \rot{veget.} & \rot{terrain} & \rot{sky} & \rot{person} & \rot{rider} & \rot{car} & \rot{truck}& \rot{bus}& \rot{train}& \rot{motorc.} & \rot{bicycle} & mIoU \\
        \midrule     
        \multirow{2}{*}{Fog}& HRNet &$95.3$&$81.1$&$89.7$&$54.2$&$48.1$&$60.3$&$76.2$&$76.0$&$89.1$&$78.1$&$98.3$&$\mathbf{58.8}$&$60.1$&$83.3$&$50.5$&$47.9$&$\mathbf{86.4}$&$49.8$&$36.3$&$69.3$ \\
        &OVeNet & $\mathbf{95.7}$&$\mathbf{82.1}$&$\mathbf{90.6}$&$\mathbf{55.5}$&$\mathbf{50.4}$&$\mathbf{63.9}$&$\mathbf{79.3}$&$\mathbf{78.2}$&$\mathbf{89.6}$&$\mathbf{78.8}$&$\mathbf{98.4}$&$56.5$&$\mathbf{63.8}$&$\mathbf{85.7}$&$\mathbf{60.3}$&$\mathbf{63.8}$&$85.5$&$\mathbf{52.7}$&$\mathbf{39.4}$ & $\mathbf{72.1}$ \\  
        \midrule
        \multirow{2}{*}{Night}&HRNet & $95.4$ & $78.1$ & $86.8$ & $46.5$ & $41.6$ & $59.1$ & $65.8$ & $65.2$ & $77.2$ & $42.0$ & $86.2$ & $66.2$ & $40.1$ & $80.5$ & $\mathbf{16.6}$ & $31.0$ & $86.2$ & $37.1$ & $49.2$ & $60.6$\\
        
        &OVeNet &$\mathbf{95.6}$&$\mathbf{79.4}$&$\mathbf{87.4}$&$\mathbf{48.5}$&$\mathbf{42.3}$&$\mathbf{61.1}$&$\mathbf{68.2}$&$\mathbf{69.9}$&$\mathbf{78.1}$&$\mathbf{43.8}$&$\mathbf{86.4}$&$\mathbf{68.6}$&$\mathbf{44.4}$&$\mathbf{82.0}$&$10.9$&$\mathbf{44.3}$&$\mathbf{87.3}$&$\mathbf{43.3}$&$\mathbf{51.3}$ & $\mathbf{62.8}$ \\
        \midrule
        \multirow{2}{*}{Rain}& HRNet &$95.7$&$ 83.9 $&$ 93.6 $&$ 60.2 $&$ 62.7 $&$ 70.1 $&$ 80.9 $&$ 79.7 $&$ 94.4 $&$ 51.9 $&$ 98.7 $&$ 72.4 $&$ 18.7 $&$ 92.5 $&$ 67.0 $&$ 85.3 $&$87.3 $&$ 50.9$&$66.4$ & $74.5$ \\
        &OVeNet & $\mathbf{96.4}$ & $\mathbf{86.5}$ & $\mathbf{94.3}$ & $\mathbf{65.2}$ & $\mathbf{64.6}$ & $\mathbf{72.6}$ & $\mathbf{83.6}$ & $\mathbf{82.2}$ & $\mathbf{94.6}$ & $\mathbf{54.3}$ & $\mathbf{98.7}$ & $\mathbf{76.4}$ & $\mathbf{21.8}$ & $\mathbf{93.4}$ & $\mathbf{75.7}$ & $\mathbf{88.9}$ & $\mathbf{89.1}$ & $\mathbf{50.9}$ & $\mathbf{66.4}$ & $\mathbf{76.6}$ \\
         \midrule
        \multirow{2}{*}{Snow}& HRNet  &$95.0 $&$ 79.6 $&$ 91.4 $&$ 49.6 $&$ 58.2 $&$ 69.7 $&$ 86.6 $&$ 78.6 $&$ 92.9 $&$ 59.4 $&$ 97.9 $&$ 77.2 $&$ 24.5 $&$ 91.7 $&$ 53.4 $&$ 56.2 $&$ 90.1 $&$ 39.8 $&$ 66.8$&$71.5$\\

        & OVeNet & $\mathbf{95.7} $&$\mathbf{81.9} $&$ \mathbf{92.4} $&$ \mathbf{55.3} $&$\mathbf{59.8} $&$ \mathbf{71.9} $&$ \mathbf{88.1} $&$ \mathbf{80.5} $&$ \mathbf{93.4} $&$ \mathbf{60.7} $&$ \mathbf{98.0} $&$ \mathbf{79.8} $&$ \mathbf{25.7} $&$ \mathbf{92.7} $&$ \mathbf{59.6} $&$ \mathbf{65.4} $&$ \mathbf{91.6} $&$ \mathbf{45.2} $&$ \mathbf{70.3}$ &$\mathbf{74.1}$\\
        \midrule
        \multirow{2}{*}{All}& HRNet & $95.3 $&$ 80.3 $&$ 90.5 $&$ 52.0 $&$ 53.1 $&$ 65.1 $&$ 78.2 $&$ 74.2 $&$ 89.2 $&$ 68.4 $&$ 96.7 $&$ 70.6 $&$ 36.1 $&$ 88.2 $&$ 55.9 $&$ 54.3 $&$ 88.0 $&$ 43.8 $&$ 58.9 $&$ 70.5$\\
        
       & OVeNet & $\mathbf{95.8} $&$ \mathbf{82.1} $&$ \mathbf{91.3} $&$ \mathbf{55.8} $&$ \mathbf{54.6} $&$ \mathbf{67.6} $&$ \mathbf{80.5} $&$ \mathbf{77.3} $&$ \mathbf{89.7} $&$ \mathbf{69.5} $&$ \mathbf{96.8} $&$ \mathbf{73.4} $&$ \mathbf{39.1} $&$ \mathbf{89.5} $&$ \mathbf{61.9} $&$ \mathbf{65.0} $&$ \mathbf{89.4} $&$ \mathbf{47.2} $&$ \mathbf{60.6} $&$ \mathbf{73.0}$ \\
        \bottomrule
    \end{tabular}
  }
    \vspace{-0.4cm}
\end{table*}

\PAR{ADE20K.} We also achieve far better results than HRNet \cite{wang2020deep} under similar training time. HRNet achieves $44.6\%$ mIoU, $80.7\%$ PixelAcc and $58.2\%$ MeanAcc while OVeNet  surpasses it, reaching $45.3\%$ mIoU, $81.3\%$ PixelAcc and $58.7\%$ MeanAcc.

\begin{figure}
  \centering
  \subfloat{\includegraphics[width=0.333\linewidth, height = 0.1\textwidth]{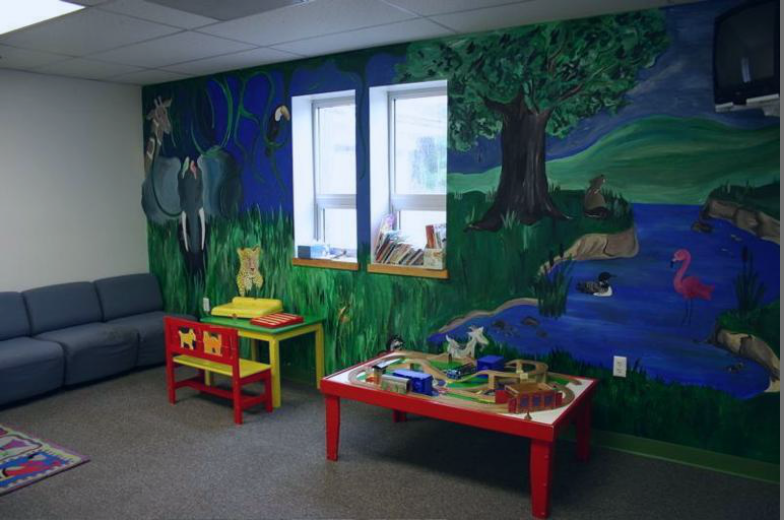}}
  \hfil
  \subfloat{\includegraphics[width=0.333\linewidth, height = 0.1\textwidth]{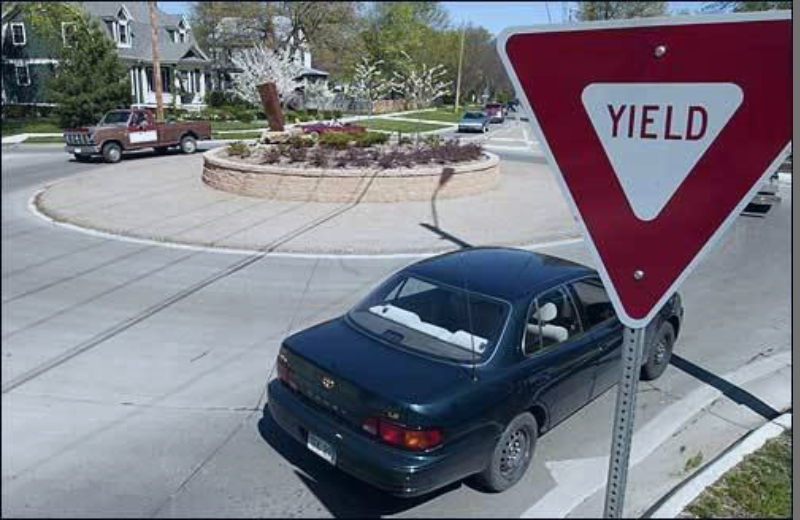}}
  \hfil
  \subfloat{\includegraphics[width=0.333\linewidth, height = 0.1\textwidth]{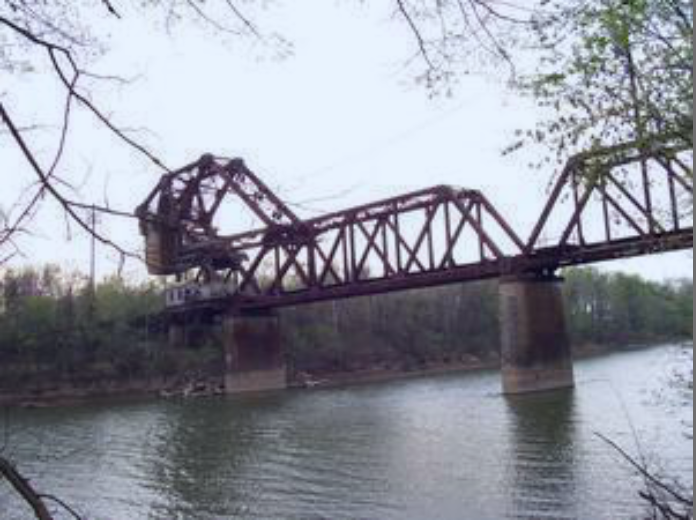}}
  
  \subfloat{\includegraphics[width=0.333\linewidth, height = 0.1\textwidth]{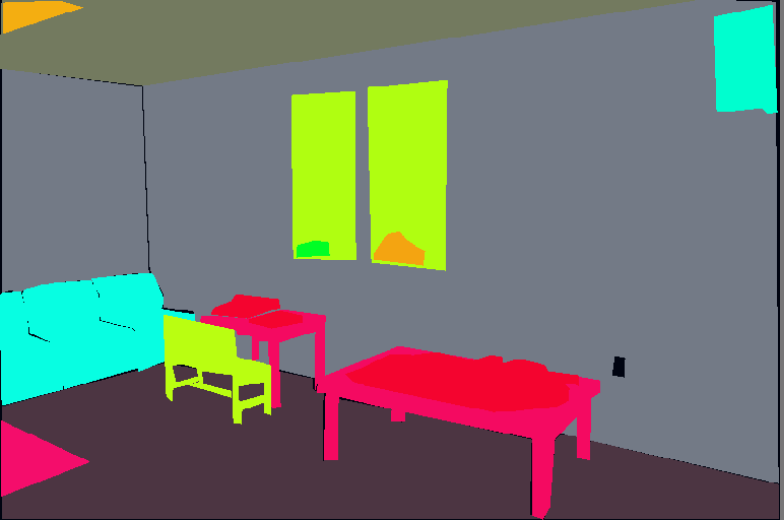}}
  \hfil
  \subfloat{\includegraphics[width=0.333\linewidth, height = 0.1\textwidth]{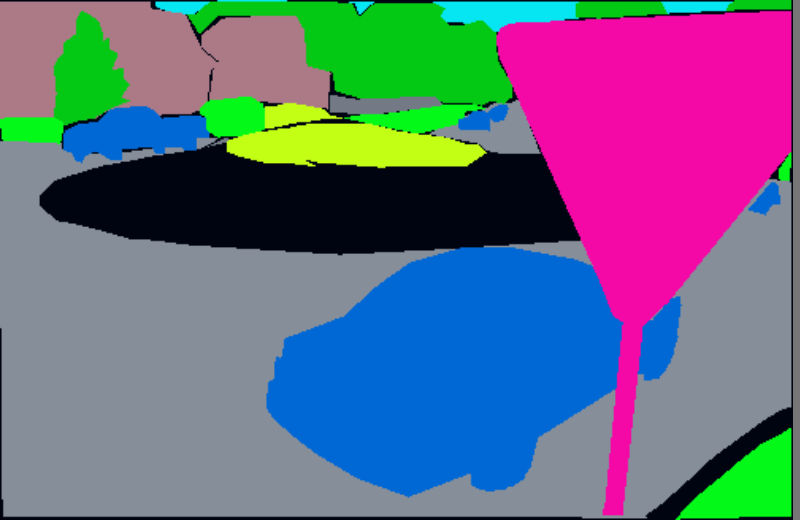}}
  \hfil
  \subfloat{\includegraphics[width=0.333\linewidth, height = 0.1\textwidth]{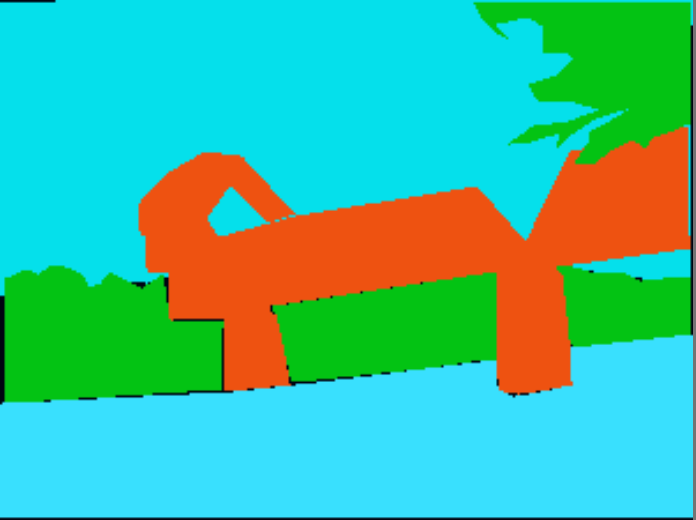}}
  
  \subfloat{\includegraphics[width=0.333\linewidth, height = 0.1\textwidth]{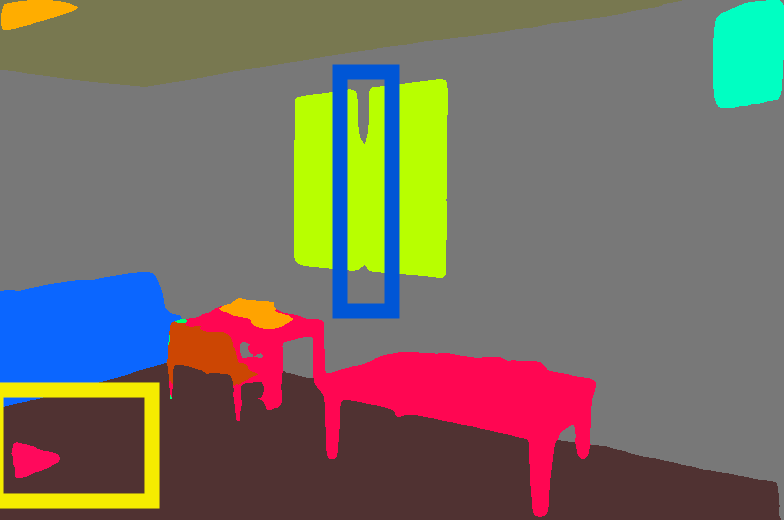}}
  \hfil
  \subfloat{\includegraphics[width=0.333\linewidth, height = 0.1\textwidth]{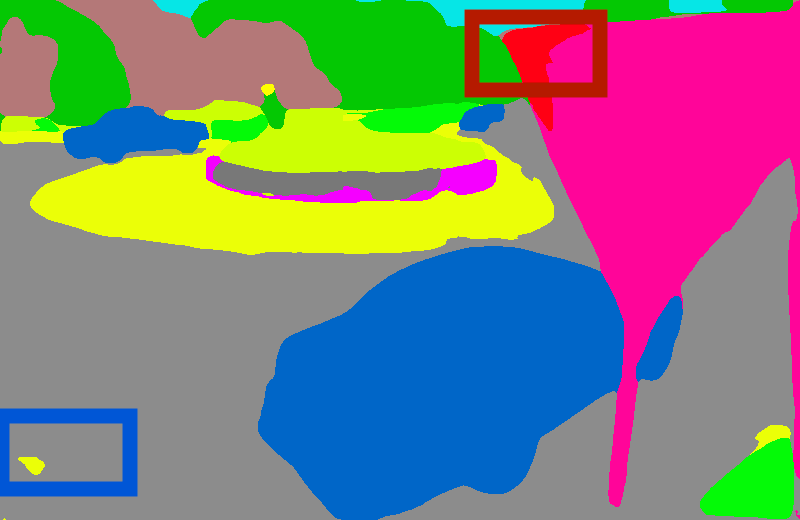}}
  \hfil
  \subfloat{\includegraphics[width=0.333\linewidth, height = 0.1\textwidth]{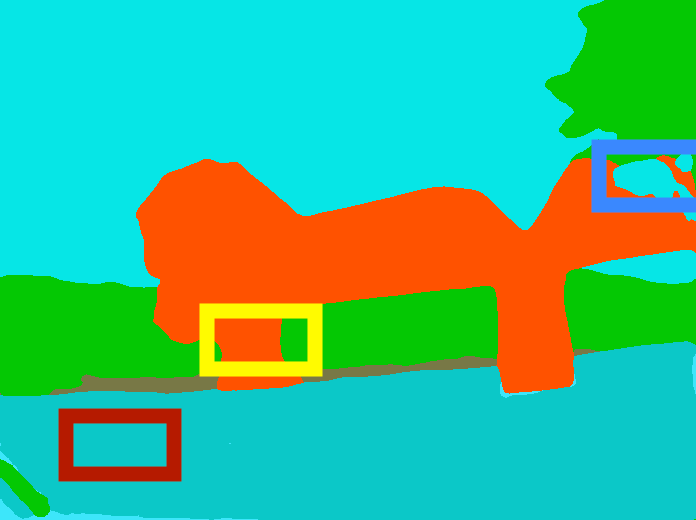}}
    
  \subfloat{\includegraphics[width=0.333\linewidth, height = 0.1\textwidth]{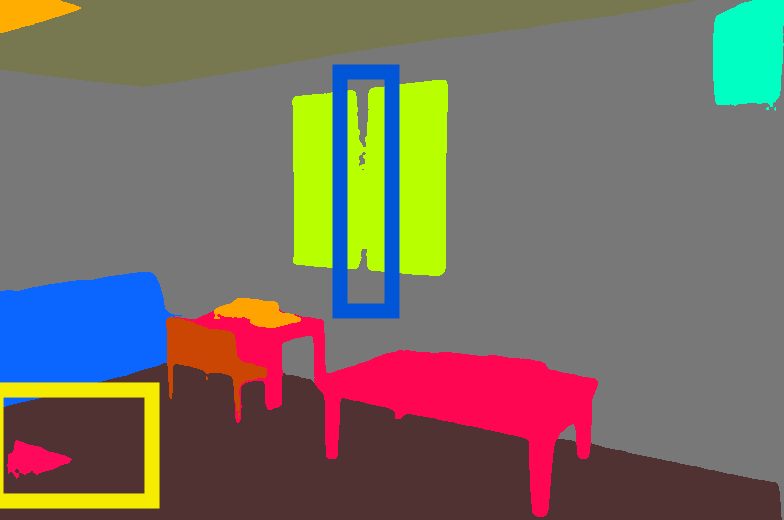}}
  \hfil
  \subfloat{\includegraphics[width=0.333\linewidth, height = 0.1\textwidth]{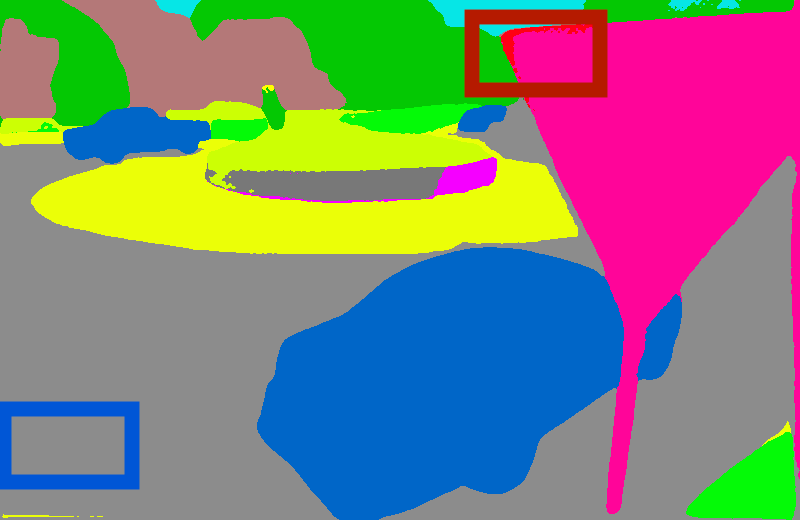}}
  \hfil
  \subfloat{\includegraphics[width=0.333\linewidth, height = 0.1\textwidth]{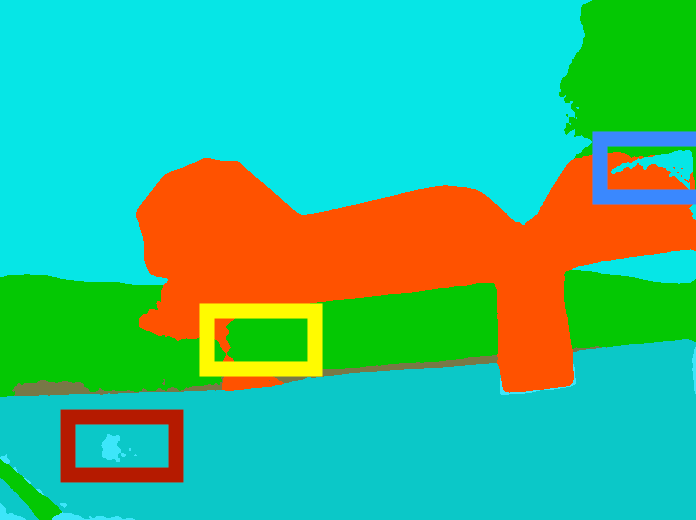}}
  \vspace*{1mm}
  \caption{ \textbf{Qualitative results of selected  examples on ADE20K}. Up to down: RGB, GT, HRNet \cite{wang2020deep}, OVeNet. Best viewed on a screen and zoomed in.}
  \label{fig:classes_ade20k}
  \vspace{-0.5cm}
\end{figure}

 Qualitative results on ADE20K support the above findings, as shown in Fig. \ref{fig:classes_ade20k}. To be more specific, in the first column, we can underline that our model tries to enlarge correctly the wall and carpet segments (blue and yellow frames). Regarding the second example, our model corrects HRNet's false prediction on the sign but also a discontinuity occurring in the road. Regarding the last example, although our offset vector-based model has some false prediction in the bridge segment (yellow frame), it corrects many erroneously classified pixels leading to a better total prediction.

\subsection{Ablation study}
\label{ablation}
In order to experimentally confirm our design choices for the offset vector-based model, we performed an ablation study, as shown  in Table \ref{fig:ablation}. We trained and evaluated 7 different variants on Cityscapes. The performance of each model variation in relation to the ground-truth images was calculated by means of the mIoU.  At first, we initialized both heads of the network with the pre-trained Imagenet weights and set the offset vector length equal to $0.5$. Secondly, we froze both main body’s and initial head’s weights. The frozen part of our model was initialized with the corresponding Cityscapes final pre-trained weights. The only part trained was the second head, which was initialized with pre-trained ImageNet weights. As shown in Table \ref{fig:ablation}, although the performance of our model is higher than the initial single-head model's one, it still remains lower than the case where both heads are trained simultaneously.  Then, we deactivated the "Frozen" feature and changed the offset vector's length logarithmically, setting it either to $1$ or $0.2$. We observed that in both cases the performance is lower than that for $0.5$. This is due to the fact that larger offset vectors point to more distant objects that may affect erroneously the final prediction, while smaller vectors do not exploit too much information from neighboring classes. Furthermore, we deactivated the OHEM Cross Entropy Loss and enabled the simple Cross Entropy Loss. As expected, the performance of the model was lower. OHEM penalizes high loss values more and leads to a better training of the model. Lastly, HRNet \cite{wang2020deep} consists of $4$ stages. In all the previous cases, branch occurred in the last ($4^{th}$) stage so as not to overload the new network with many extra parameters. When branching in the $3^{rd}$ stage, the performance did not improve.

\begin{table}
 \centering
  \caption{\textbf{Comparison of the models on different conditions of ACDC.}}
 \label{tab:conditions}
 \begin{tabular}{lccccc}
\toprule
\textbf{Method} & \textbf{Fog}&\textbf{Night}& \textbf{Rain}&\textbf{Snow} & \textbf{All} \\
\midrule
RefineNet \cite{lin2017refinenet} & $65.7$ &$55.5$&$68.7$&$65.9$& $65.3$\\
DeepLabv2 \cite{chen2017deeplab} & $54.5$ &$45.3$&$59.3$&$57.1$& $55.3$\\
DeepLabv3+ \cite{chen2018encoder} & $69.1$ &$60.9$&$74.1$&$69.6$& $70.0$\\
HRNet  \cite{wang2020deep} & $69.3$ &$60.6$&$74.5$&$71.5$& $70.5$\\
\midrule
OVeNet & $\mathbf{72.1}$ &$\mathbf{62.8}$&$\mathbf{76.6}$&$\mathbf{74.1}$& $\mathbf{73.0}$\\
 \bottomrule
 \end{tabular}
   \vspace{-0.1cm}
\end{table}

\begin{table}
 \centering
  \caption{\textbf{Ablation study of components of our method.} ``\textbf{Fr}": Frozen main body’s and initial head’s weights initialized with HRNet's final Cityscapes weights,   ``\textbf{Br}": Branch, ``$\boldsymbol{\tau}$": Offset Vector Length, ``\textbf{OHEM}": OHEM Cross Entropy.}
 \label{fig:ablation}
 \begin{tabular}{cccc|c}
\toprule
\textbf{Fr}  & \textbf{Br} & $\boldsymbol{\tau}$ & \textbf{OHEM} & \textbf{mIoU} \\
\midrule
 &    &   & \checkmark & $81.83$ \\
 &   $4$ & $0.5$ &   \checkmark & $\mathbf{82.40}$ \\
 \checkmark &  $4$ & $0.5$ &  \checkmark & $82.01$ \\
   &    $4$ & $1 $&  \checkmark & $81.79$ \\
    &    $4$ & $0.2$ &  \checkmark & $82.30$ \\
    &    $4$ & $0.2$ & &   $81.96$ \\
    &   $3$ & $0.2$ &\checkmark &   $82.20$ \\
 \bottomrule
 \end{tabular}
   \vspace{-0.6cm}
\end{table}

\section{Conclusion}
All in all, we have presented OVeNet, a supervised model for semantic segmentation, which selectively exploits information from neighboring pixels to improve initial semantic predictions. OVeNet excels both in global and per-class performance across most classes on three widely used semantic segmentation benchmarks. By correcting misclassified pixels, it reduces discontinuities and improves the shapes of segments, leading to more realistic results. This is a highly relevant contribution for real-world applications that depend on semantic segmentation, such as autonomous cars or medical imaging and diagnostics.

{\small
\bibliographystyle{ieee_fullname}
\bibliography{egbib}
}
\clearpage
\appendix

\section{Network Instantiation}

OVeNet follows the same general architecture as~\cite{wang2020deep}. Since it is built on HRNet~\cite{wang2020deep}, our network contains four stages and two heads, as shown in Table~\ref{tab:hrnet-block}. The branch occurs on the 4$^{\text{th}}$ stage. The semantic head of our model is constructed from modularized blocks, which are repeated a specific number of times in each of the four stages. In particular, the blocks are repeated $1$, $1$, $4$, and $3$ times, respectively, following the same configuration as the original HRNet model. However, in the offset head, we had to make a modification to the number of block repetitions in order to address memory constraints. As a result, the blocks are repeated 2 times in the 4$^{\text{th}}$ stage of the offset head. Each modularized block in our network consists of $1$, $2$, $3$, or $4$ branches, depending on the stage in which it is located (1$^{\text{st}}$, 2$^{\text{nd}}$, 3$^{\text{rd}}$, or 4$^{\text{th}}$). Each branch is associated with a different resolution and is composed of four residual units and one multi-resolution fusion unit.
\newcommand{\blocka}[3]{\multirow{3}{*}{
 $\left[\begin{array}{c}{3\times3, #1}\\[-.1em] {3\times3, #1} \end{array}
\right]
\times#2 
\times#3$
}
}

\newcommand{\blockb}[3]{\multirow{3}{*}{\(\left[\begin{array}{c}{1\times1, #2}\\[-.1em] {3\times3, #2}\\[-.1em] {1\times1, #1}\end{array}\right]\)
$\times$#3}
}

\renewcommand{\arraystretch}{1.3}
\begin{table*}[!htbp]
\setlength{\tabcolsep}{13pt}
\scriptsize
\caption{
\footnotesize 
\textbf{The architecture of the OVeNet (main body)}. Within each cell of our network, there are three distinct components. The first component, represented by $[\cdot]$, refers to the residual unit. The second component is a numerical value that specifies the number of times the residual unit is repeated. The final component is another numerical value that indicates how many times the modularized block is repeated within the cell. In each residual unit, the variable $C$ is used to represent the number of channels.
}
\label{tab:hrnet-block}
\vspace*{1.5mm}
\centering
\resizebox{\linewidth}{!}{
	\begin{tabular}{c|c|cccc}
	\toprule
	 Head & Resolution & Stage $1$ & Stage $2$ & Stage $3$ & Stage $4$ \\
    \midrule
    \multirow{12}{*}{Semantic} 
	&\multirow{3}{*}{$4\times$} 
	& \blockb{256}{64}{$4\times1$} & \blocka{C}{4}{1}  & \blocka{C}{4}{4} & \blocka{C}{4}{3} \\
	&&  &  &  &  \\
	&&  &  &  &  \\
	&\multirow{3}{*}{$8\times$}
	&  & \blocka{2C}{4}{1}  & \blocka{2C}{4}{4} & \blocka{2C}{4}{3}\\
	&&  &  &  &  \\
	&&  &  &  &  \\
	&\multirow{3}{*}{$16\times$}
	&  &  & \blocka{4C}{4}{4} & \blocka{4C}{4}{3} \\
	&&  &  &  &  \\
	&&  &  &  &  \\
	&\multirow{3}{*}{$32\times$}
	&  &  &  & \blocka{8C}{4}{3}\\
	&&  &  &  &  \\
	&&  &  &  &  \\
	\midrule
        \multirow{12}{*}{Offset Vector } 
	&\multirow{3}{*}{$4\times$} 
	&  &  &  & \blocka{C}{4}{2} \\
	&&  &  &  &  \\
	&&  &  &  &  \\
	&\multirow{3}{*}{$8\times$}
	&  &   &  & \blocka{2C}{4}{2}\\
	&&  &  &  &  \\
	&&  &  &  &  \\
	&\multirow{3}{*}{$16\times$}
	&  &  &  & \blocka{4C}{4}{2} \\
	&&  &  &  &  \\
	&&  &  &  &  \\
	&\multirow{3}{*}{$32\times$}
	&  &  &  & \blocka{8C}{4}{2}\\
	&&  &  &  &  \\
	&&  &  &  &  \\
	\bottomrule
	\end{tabular}
 }
\end{table*}

\section{Additional Qualitative Results}

\begin{figure*}
  \centering
  \subfloat{\includegraphics[width=0.95\linewidth]{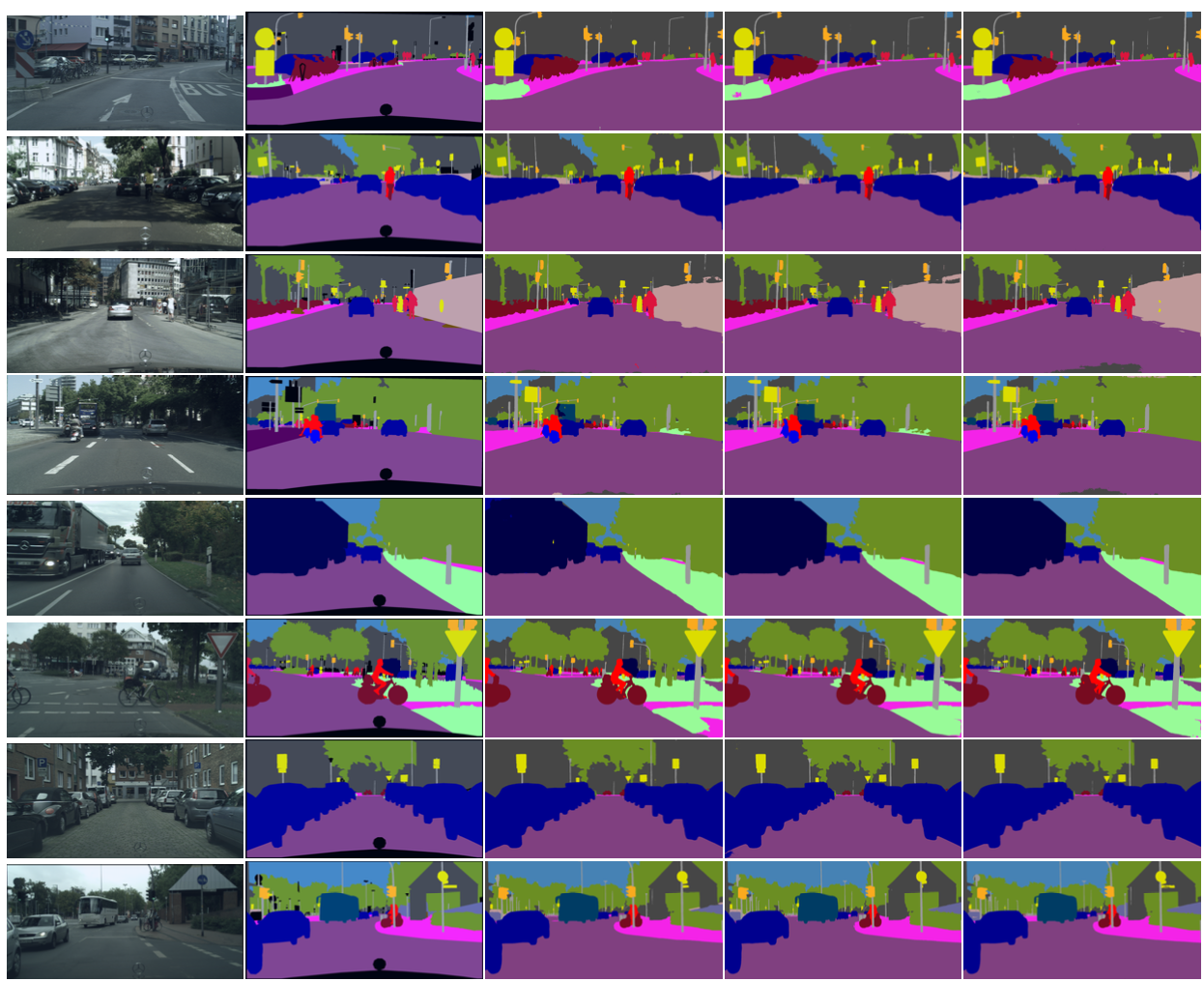}}
   \hfil
  \vspace*{1mm}
  \caption{\textbf{Additional qualitative results of selected  examples on Cityscapes}. From left to right: input image, ground-truth annotation, and prediction with HRNet~\cite{wang2020deep}, OVeNet (\emph{HRNet}), and OVeNet (\emph{HRNet$+$OCR}). Best viewed on a screen and zoomed in.}
  \label{fig:true_comp_city}
  \vspace*{1mm}
\end{figure*}

\begin{figure*}
  \centering
  \subfloat{\includegraphics[width=0.95\linewidth]{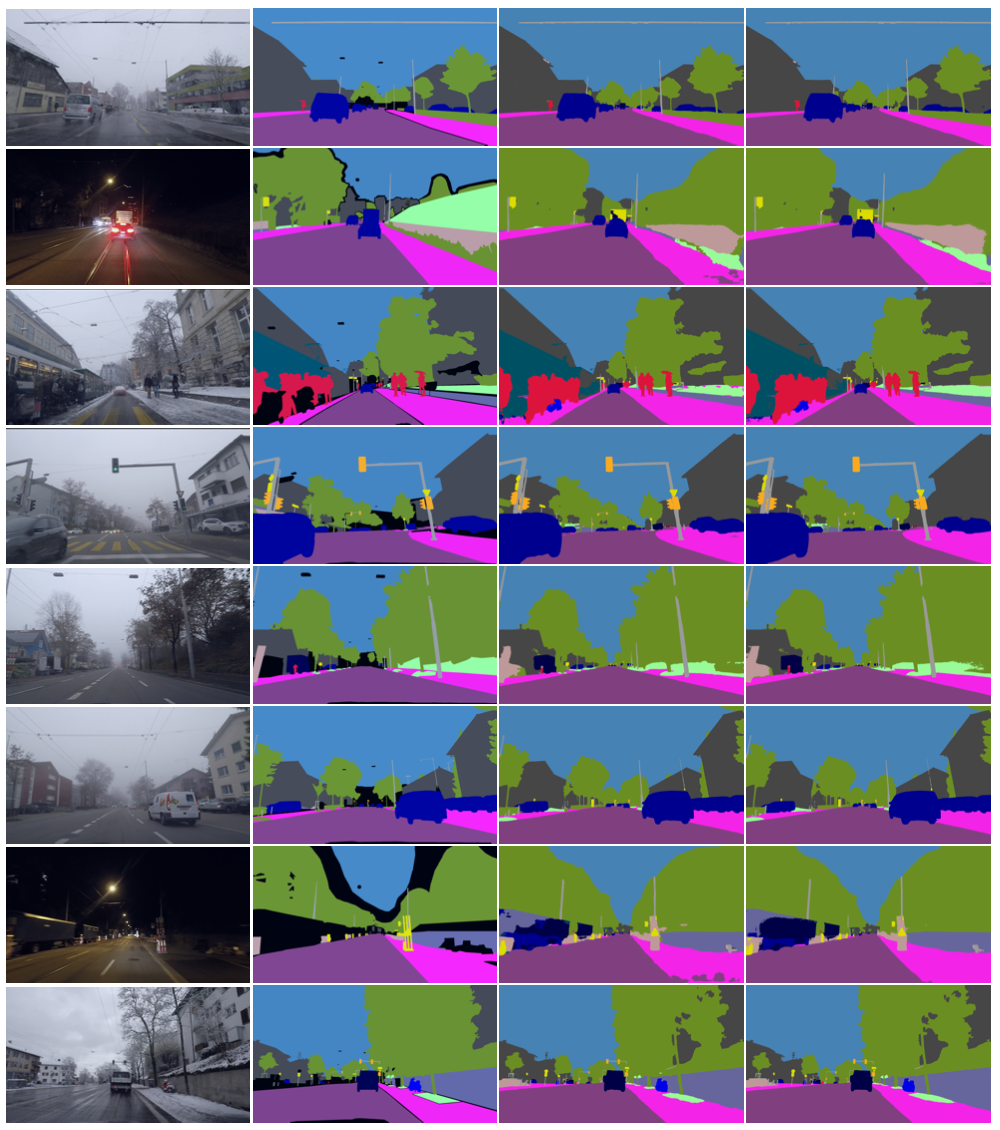}}
  \hfil
  \vspace*{1mm}
  \caption{\textbf{Additional qualitative results of selected  examples on ACDC}. From left to right: input image, ground-truth annotation, and prediction with HRNet~\cite{wang2020deep} and OVeNet. Best viewed on a screen and zoomed in.}
  \label{fig:true_comp_acdc}
  \vspace*{1mm}
\end{figure*}

\begin{figure*}
  \centering
  \subfloat{\includegraphics[width=0.95\linewidth]{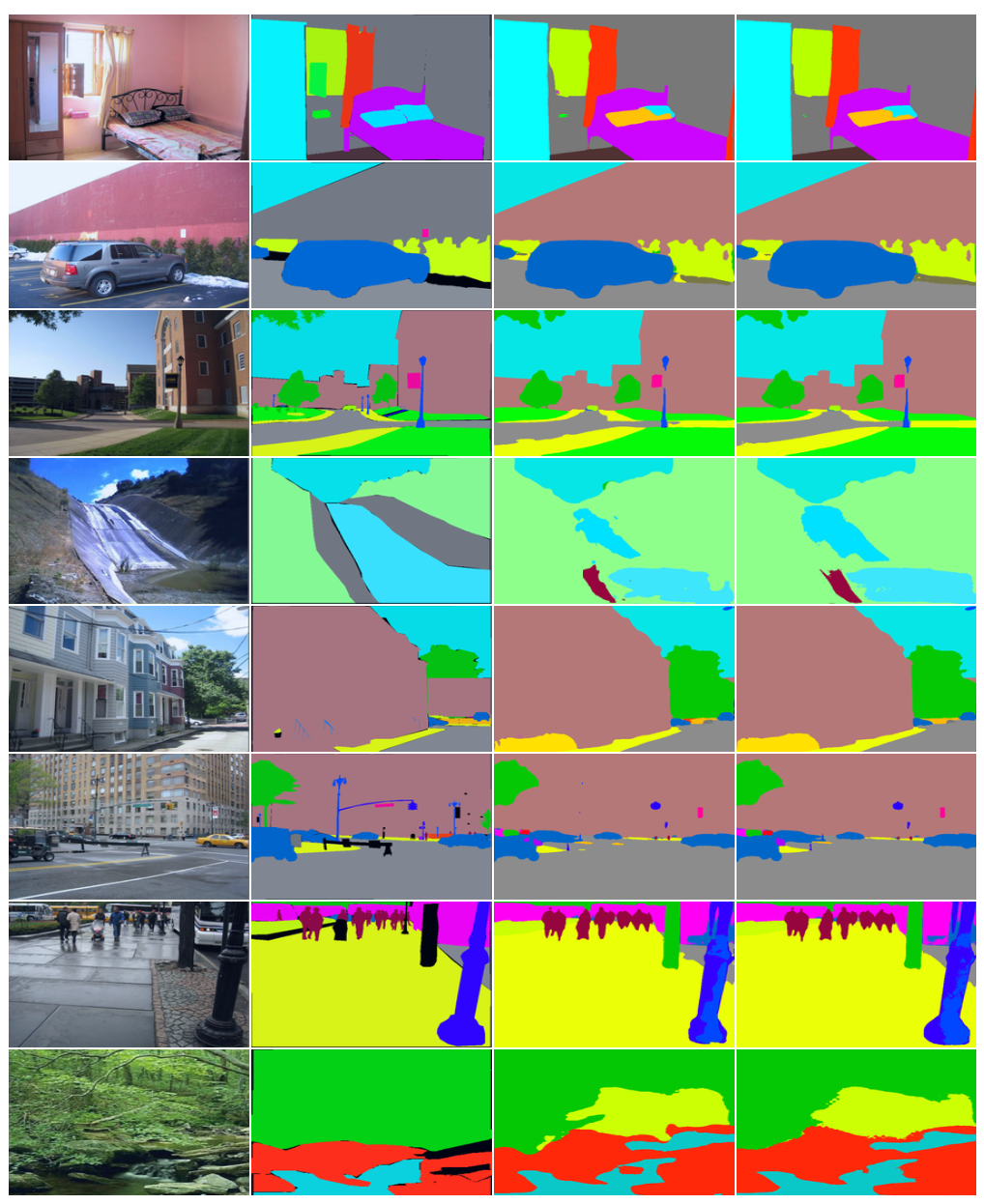}}
  \hfil
  \vspace*{1mm}
  \caption{\textbf{Additional qualitative results of selected  examples on ADE20K}. From left to right: input image, ground-truth annotation, and prediction with HRNet~\cite{wang2020deep} and OVeNet. Best viewed on a screen and zoomed in.}
  \label{fig:true_comp_ade20k}
  \vspace*{1mm}
\end{figure*}

\begin{figure*}
  \centering
 \subfloat{\includegraphics[width=0.95\linewidth]{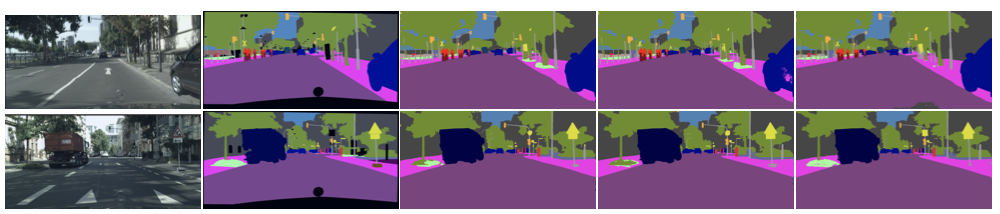}}
  \vspace*{1mm}
  \caption{\textbf{Challenging cases on Cityscapes}. From left to right: input image, ground-truth annotation, and prediction with HRNet~\cite{wang2020deep}, OVeNet (\emph{HRNet}), and OVeNet (\emph{HRNet$+$OCR}).
  Best viewed on a screen and zoomed in.}
  \label{fig:false_comp_city}
  \vspace*{1mm}
\end{figure*}

\begin{figure*}
  \centering
    \subfloat{\includegraphics[width=0.95\linewidth]{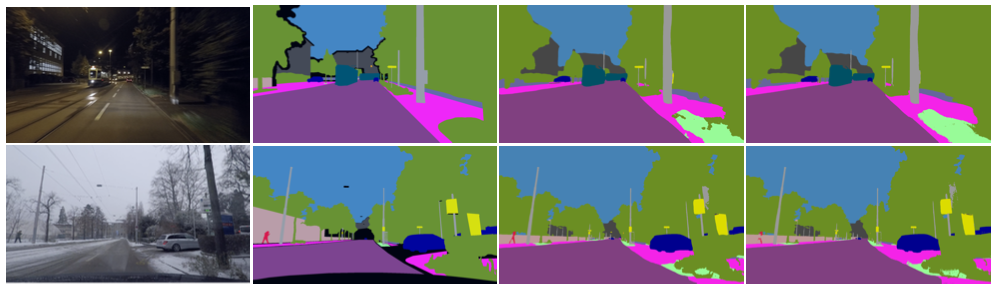}}
  \vspace*{1mm}
  \caption{\textbf{Failure cases on ACDC}. From left to right: input image, ground-truth annotation, and prediction with HRNet~\cite{wang2020deep} and OVeNet. Best viewed on a screen and zoomed in.}
  \label{fig:false_comp_acdc}
  \vspace*{1mm}
\end{figure*}

\begin{figure*}
  \centering
    \subfloat{\includegraphics[width=0.95\linewidth]{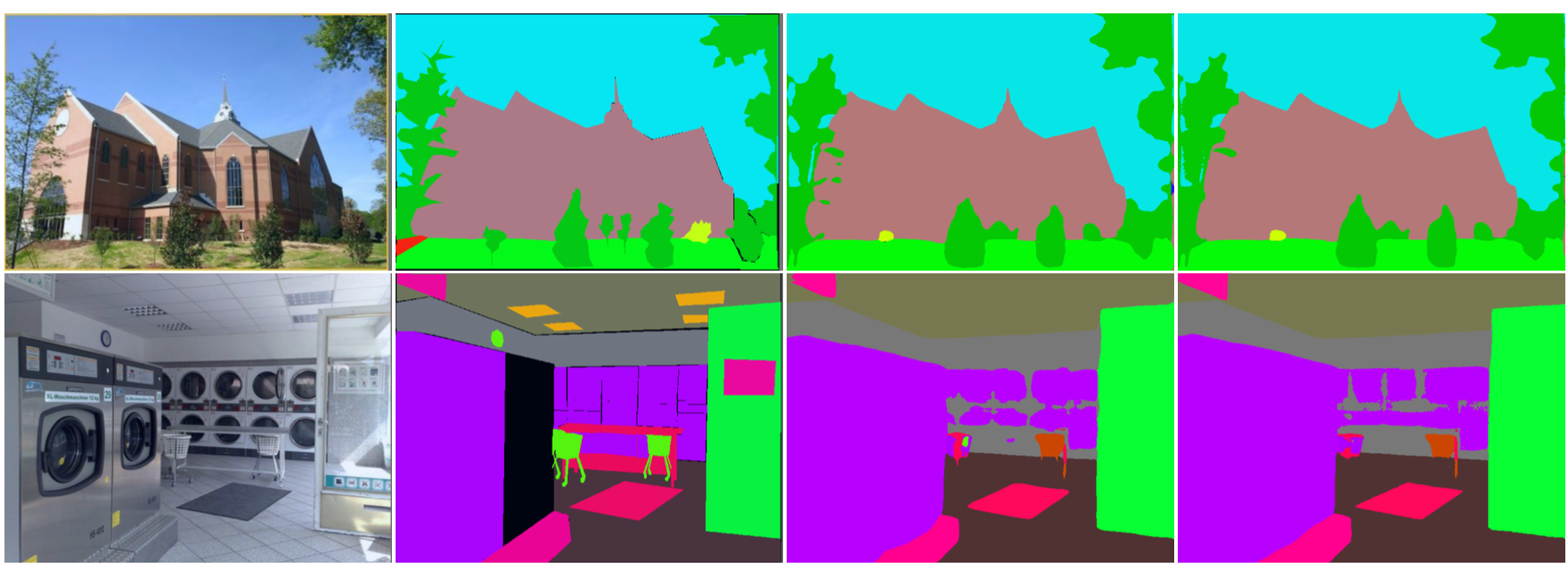}}
  \vspace*{1mm}
  \caption{\textbf{Failure cases on ADE20K}. From left to right: input image, ground-truth annotation, and prediction with HRNet~\cite{wang2020deep} and OVeNet. Best viewed on a screen and zoomed in.}
  \label{fig:false_comp_ade20k}
  \vspace*{1mm}
\end{figure*}

We provide additional qualitative comparisons of OVeNet to its HRNet baseline for the three examined datasets: Cityscapes~\cite{cordts2016cityscapes}, ACDC~\cite{sakaridis2021acdc} and ADE20K \cite{zhou2016semantic}. More specifically, we provide sets of successful segmentations in Fig.~\ref{fig:true_comp_city}, \ref{fig:true_comp_acdc} and \ref{fig:true_comp_ade20k} and sets of challenging and failure cases in Fig.~\ref{fig:false_comp_city}, \ref{fig:false_comp_acdc} and \ref{fig:false_comp_ade20k} correspondingly. 

In Fig.~\ref{fig:true_comp_city} we depict the successful results on Cityscapes. We also present an additional comparison of the default instance of OVeNet built only on HRNet~\cite{wang2020deep} with the instance of OVeNet built on HRNet$+$OCR~\cite{YuanCW18}. As we can see, OVeNet (\emph{HRNet})  reduces correctly the number of misclassified terrain pixels on the right side of the road in the $3^{rd}$ row of Fig. ~\ref{fig:true_comp_city}. On the other hand, OVeNet (\emph{HRNet $+$ OCR}) achieves a better prediction since it enlarges the sidewalk segment and it eliminates the terrain. Regarding the $6^{th}$ row, OVeNet (\emph{HRNet}) expands the terrain on the right, while the HRNet $+$ OCR-based model increases additionally the sidewalk segment in the background. Overall, the latter generally achieves better predictions than the former.

In Fig. \ref{fig:true_comp_acdc} we observe the results of successful segmentations on ACDC, where  HRNet~\cite{wang2020deep} is compared to OVeNet built only on it. Specifically, in the $2^{nd}$ as well as the $7^{th}$ row of Fig. \ref{fig:true_comp_acdc}, OVeNet enhances the prediction made by HRNet in the sidewalk on the right. A similar result happens with the terrain segment in the $5^{th}$ row, as it covers correctly all the more space. Thus, our model surpasses baseline's performance on adverse conditions. 

In Fig. \ref{fig:true_comp_ade20k} we depict the results of successful segmentations on ADE20K. To be more specific, in the $3^{rd}$ as well as the $4^{th}$ row, OVeNet enhances the prediction made by HRNet by enlarging some segments (river and house respectively). As a result, our model surpasses baseline's performance on everyday images.

As for Fig.~\ref{fig:false_comp_city}, we  provide some challenging cases on Cityscapes. To be more specific, we can see that in both rows, HRNet~\cite{wang2020deep} results in a better prediction than OVeNet (\emph{HRNet}) (e.g car on the left in the $1^{st}$ row, terrain on the right in the $2^{nd}$ one). On the other hand, OVeNet (\emph{HRNet $+$ OCR}) outperforms both the HRNet and HRNet-based model, leading to a better total outcome.

Regarding Fig. \ref{fig:false_comp_acdc}, we see some some failure cases of our model built on HRNet~\cite{wang2020deep} on ACDC. Specifically, in both rows, there are more correctly predicted labels in the vegetation segment on the right outputted by  HRNet than by OVeNet.

As for Fig. \ref{fig:false_comp_ade20k}, we see some some false results predicted our model built on HRNet~\cite{wang2020deep} on ADE20K. Specifically, in both set of images, there are more correctly predicted labels in the tree and wash machine segment respectively outputted by HRNet than by OVeNet.

All in all, we observe that OVeNet not only produces results that are very faithful to ground-truth annotations, but its predictions also surpass the predictions made by the initial HRNet in terms of quality. By correctly classifying several pixels which are misclassified by the HRNet baselines, OVeNet (\emph{HRNet $+$ OCR}) eliminates inconsistencies and enhances the shape and appearance of respective segments, resulting in more realistic outputs.

 \newpage

\newpage

\end{document}